\documentclass[10pt,twocolumn,letterpaper]{article}

\usepackage[pagenumbers]{cvpr} %

\makeatletter
\renewcommand\paragraph{\@startsection{paragraph}{4}{\z@}%
	{0.7ex \@plus.3ex \@minus.2ex}%
	{-1em}%
	{\normalfont\normalsize\bfseries\maybe@addperiod}}
\newcommand{\maybe@addperiod}[1]{#1\@addpunct{.}}
\makeatother

\definecolor{cvprblue}{rgb}{0.21,0.49,0.74}
\usepackage{colortbl}
\usepackage{xcolor}
\usepackage[pagebackref,breaklinks,colorlinks,allcolors=cvprblue]{hyperref}
\usepackage{multirow}
\usepackage{booktabs}
\usepackage{tabularx}
\usepackage{arydshln}
\usepackage{float}
\usepackage{graphicx}
\usepackage{cuted}
\usepackage{pifont}
\usepackage{listings}
\usepackage{algorithmic}
\usepackage{algorithm}
\usepackage{bm} 
\usepackage{amssymb}
\usepackage{amsmath}
\usepackage{etoolbox}
\usepackage{longtable}
\usepackage{hyperref}
\makeatletter
\AfterEndEnvironment{algorithm}{\let\@algcomment\relax}
\AtEndEnvironment{algorithm}{\kern2pt\hrule\relax\vskip3pt\@algcomment}
\let\@algcomment\relax
\newcommand\algcomment[1]{\def\@algcomment{\footnotesize#1}}
\renewcommand\fs@ruled{\def\@fs@cfont{\bfseries}\let\@fs@capt\floatc@ruled
  \def\@fs@pre{\hrule height.8pt depth0pt \kern2pt}%
  \def\@fs@post{}%
  \def\@fs@mid{\kern2pt\hrule\kern2pt}%
  \let\@fs@iftopcapt\iftrue}
\makeatother

\definecolor{winered}{RGB}{115, 0, 57} 
\makeatletter
\def\adl@drawiv#1#2#3{%
        \hskip.5\tabcolsep
        \xleaders#3{#2.5\@tempdimb #1{1}#2.5\@tempdimb}%
                #2\z@ plus1fil minus1fil\relax
        \hskip.5\tabcolsep}
\newcommand{\cdashlinelr}[1]{%
  \noalign{\vskip\aboverulesep
           \global\let\@dashdrawstore\adl@draw
           \global\let\adl@draw\adl@drawiv}
  \cdashline{#1}
  \noalign{\global\let\adl@draw\@dashdrawstore
           \vskip\belowrulesep}}
\makeatother

\def\paperID{2203} %
\def\confName{CVPR}
\def\confYear{2025}

\title{GPS as a Control Signal for Image Generation}
\author{Chao Feng\textsuperscript{1}  \quad
Ziyang Chen\textsuperscript{1} \quad
Aleksander Ho{\l}y{\'n}ski\textsuperscript{2} \quad
Alexei A. Efros\textsuperscript{2} \quad
Andrew Owens\textsuperscript{1} 
\vspace{3mm} \\
\textsuperscript{1}University of Michigan \quad \textsuperscript{2}UC Berkeley \vspace{1.25mm} \\
{\normalsize \texttt{\url{https://cfeng16.github.io/gps-gen/}}}
}
\begin{document}
\maketitle

\begin{strip}
    \centering
    \vspace{-12mm}
     \includegraphics[width=\linewidth]{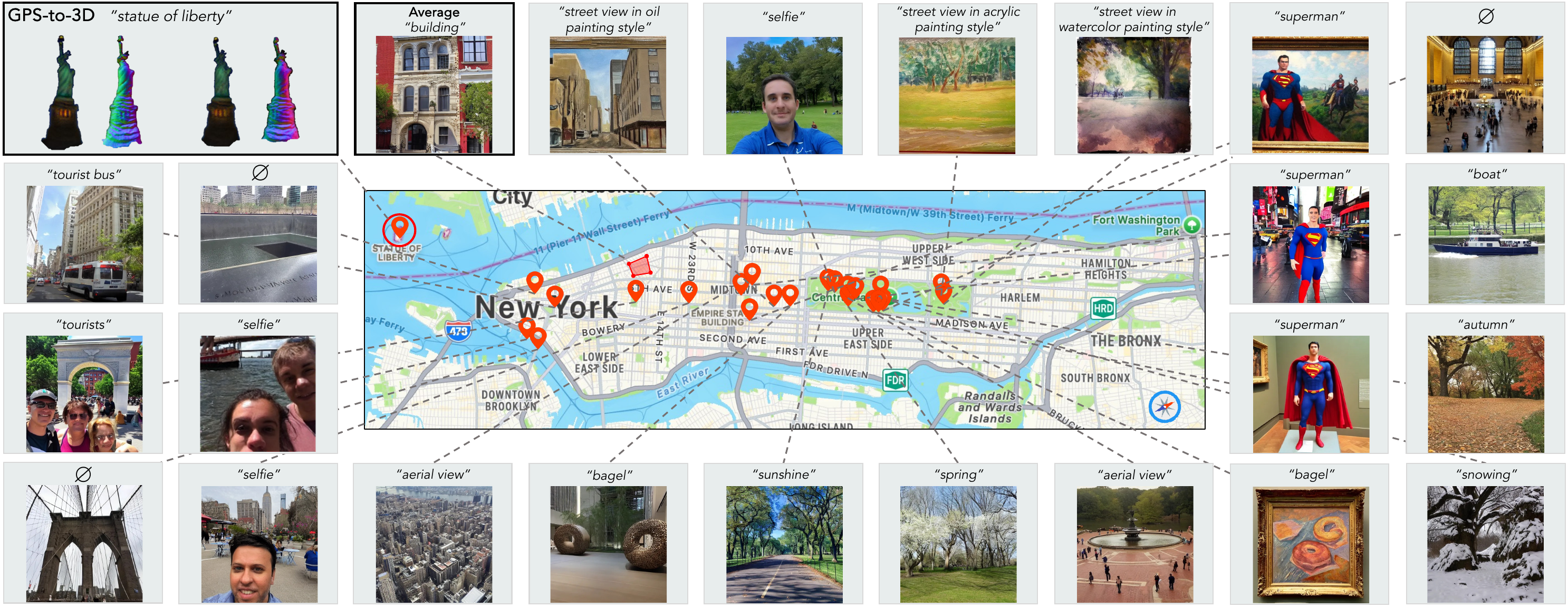}\vspace{2mm}
    \captionof{figure}{{\bf What can we do with a GPS-conditioned image generation model?} 
    We train GPS-to-image models and use them for tasks that require a fine-grained understanding of how images vary within a city. For example, a model trained on densely sampled geotagged photos from Manhattan can generate images that match a neighborhood's general appearance and capture key landmarks like museums and parks. We show images sampled from a variety of GPS locations and text prompts. For example, an image with the text prompt ``bagel'' results in a modern-style sculpture when conditioned on the Museum of Modern Art and an impressionist-style painting when conditioned on the Metropolitan Museum of Art. We also ``lift'' a 3D NeRF of the Statue of Liberty from a landmark-specific 2D GPS-to-image model using score distillation sampling. {\bf Please see the \href{https://cfeng16.github.io/gps-gen/}{project webpage} and Sec.~\ref{supp:more_qua_results} for more examples.}}
    \vspace{-3mm}
    \label{fig:teaser}

\end{strip}

\begin{abstract}
We show that the GPS tags contained in photo metadata provide a useful control signal for image generation. We train GPS-to-image models and use them for tasks that require a fine-grained understanding of how images vary within a city. In particular, we train a diffusion model to generate images conditioned on both GPS and text. The learned model generates images that capture the distinctive appearance of different neighborhoods, parks, and landmarks. We also extract 3D models from 2D GPS-to-image models through score distillation sampling, using GPS conditioning to constrain the appearance of the reconstruction from each viewpoint. Our evaluations suggest that our GPS-conditioned models successfully learn to generate images that vary based on location, and that GPS conditioning improves estimated 3D structure.

\end{abstract}
\vspace{-5mm}
 
\section{Introduction}
\label{sec:intro}

Each time a tourist snaps a photo, they capture a tiny sliver of the world. Research on geotagged photo collections has shown that these images, when analyzed collectively, can reveal surprising amounts of information, including the landmarks that people visit~\cite{crandall2009mapping}, the 3D structure of the buildings they see~\cite{snavely2006photo}, and geographic variations in architecture and fashion~\cite{doersch2012makes,mall2019geostyle}.

In this paper, we show that GPS conditioning is a useful and abundantly available control signal for image generation, which complements other common forms of conditioning like text.  We train diffusion models to map GPS coordinates from a particular city (or from a more spatially localized region) to images. To solve this problem, the model needs to capture fine-grained distinctions in how images change their appearance over space. Such a model, for example, needs to know the locations of museums and parks, the subtle differences between building facades in different neighborhoods, and how landmarks change their appearance from different perspectives. Consequently, these models convey information that would be difficult to obtain from image or language supervision alone.

We demonstrate the utility of this location-based control signal in a variety of ways. First, we train diffusion models on both GPS coordinates and text (obtained from captioning), allowing us to generate images that appear as though they were shot in a given location while capturing a particular text prompt to allow for additional control (\cref{fig:teaser}). 
The resulting model exhibits the ability to perform compositional generation and the ability to closely follow location conditioning. For example, the prompt ``aerial view'' produces a plausible overhead image of  Central Park's Bethesda Fountain. The prompt ``superman'' results in a statue or a painting when conditioned on locations within the New York Museum of Modern Art, while it generates a photo of a costumed human in Times Square. 

Second, we show that 3D geometry can be lifted from 2D GPS-to-image models (Fig.~\ref{fig:teaser}, Statue of Liberty). We exploit the fact that GPS conditioning tells us how a landmark should appear from different viewing positions. Given a GPS-to-photo model trained on a specific landmark, we extract a NeRF using score distillation sampling~\cite{poole2023dreamfusion,wang2023score}, using the learned conditional distribution to ensure that the estimated NeRF is consistent with the visual appearance of photos from every viewing direction. This ``3D reconstruction by 3D generation'' approach does not require explicit camera pose estimation, matching, or triangulation. Instead, it obtains its signal from cross-modal association between GPS and images.

Our evaluations suggest that GPS conditioning provides a useful control signal for generating images and extracting structure from geotagged image collections. These experiments suggest:
\def\labelitemi{\textbullet }
\begin{itemize}[leftmargin=*,topsep=1pt, noitemsep]
\item GPS-conditioned image generation models can capture subtle variations between locations within a city. 
\item GPS conditioning complements language-based conditioning for image generation and 3D generation.
\item 3D reconstructions can be extracted from 2D GPS-to-image models %
without explicit camera pose estimation.
\end{itemize}

\vspace{-2mm} \section{Related Work}
\label{related_work}

\paragraph{Exploring geotagged image collections}
GPS tags associated with images have been used in many different ways. 
Some researchers use GPS coordinates as complementary signals for image classification~\cite{tang2015improving} and remote sensing~\cite{christie2018functional}. 
Some works predict geolocation from images~\cite{haas2024pigeon,hays2008im2gps,weyand2016planet,suwono2023location,zhang2024geodtr+} or retrieve GPS or address from images in CLIP style~\cite{vivanco2024geoclip, xu2024addressclip}.
Other work has applied GPS data for different applications, such as city mapping and landmark identification~\cite{crandall2009mapping}, architectural styles~\cite{doersch2012makes}, scene chronology~\cite{matzen2014scene}, and fashion trends~\cite{mall2019geostyle}. \citet{mall2022discovering} creates ``underground map'' of cities by analyzing fashion styles in public social media photos to reveal unique neighborhood information. \citet{snavely2006photo} used geo-tagged, unordered photos to reconstruct 3D models of tourist sites and formulated it as a photo tourism problem.
\citet{shrivastava2011data} introduced the {\em painting2GPS} task, which estimates the GPS coordinates of a painting by matching it to a collection of real geo-tagged photos. In contrast to previous works, we leverage GPS tags as additional conditional signals in generative models, enabling GPS-guided generation of tourist images and providing free supervisory signals for 3D reconstruction.

\paragraph{Conditioning diffusion models.} Diffusion models~\cite{ho2020denoising,song2020denoising,song2020score,song2019generative,dhariwal2021diffusion,rombach2022high,peebles2023scalable,saharia2022photorealistic,blattmann2023stable,ramesh2022hierarchical,sohl2015deep} are designed to learn how to restore data that has been deliberately corrupted by adding Gaussian noise. Specifically, the forward process gradually adds noise to data over several time steps, transforming it into pure noise. The reverse process then learns to denoise the data step by step, reconstructing the original data from the noise. 
Prior work has used many various conditions to guide diffusion models for image/video/3D/4D synthesis, including text~\cite{saharia2022photorealistic,ramesh2022hierarchical,blattmann2023stable,rombach2022high,brooks2023instructpix2pix,poole2023dreamfusion,bahmani20244d,chen2024videocrafter2,lin2023magic3d,tang2024codi,tang2024any}, depth~\cite{zhang2023adding,cai2024generative}, audio~\cite{biner2024sonicdiffusion,tang2024any,tang2024codi,girdhar2023imagebind}, camera poses~\cite{shi2023mvdream,kuang2024collaborative,bahmani2024vd3d,kumari2024customizing,liu2023zero,sargent2023zeronvs}, motion~\cite{shi2024dragdiffusion,wang2024motionctrl,yin2023dragnuwa,li2024dragapart,bahmani2025tc4d,geng2024motion}, tactile signals~\cite{yang2023generating,yang2024binding}, segmentation mask~\cite{bar2023multidiffusion,zhang2023adding,bahmani2023cc3d}, and many other conditions. \citet{siglidis2024diffusion} utilizes conditional diffusion models as data mining tools. Recently, \citet{deng2024streetscapes} uses street maps, height maps, and text as conditions for video generative models to synthesize streetscapes. \citet{khanna2023diffusionsat} generate satellite images from GPS coordinates but are limited to satellite imagery and require specialized, calibrated training data.l In contrast, our work learns from ``in the wild'' geotagged photos using the GPS tags taken from EXIF metadata, a much more diverse and abundantly available data source, and we use our models for a variety of downstream tasks that were not considered in prior work, such as 3D model extraction.

\paragraph{3D reconstruction and generation}
Reconstructing 3D models from multiple images~\cite{hartley2003multiple} is a longstanding problem. The traditional pipeline involves matching and verifying features~\cite{lowe2004distinctive,nister2006scalable}, estimating camera pose and sparse 3D geometry with structure from motion (SfM)~\cite{tomasi1992shape,wu2011multicore,schonberger2016structure}, and generating dense reconstructions using multi-view stereo~\cite{furukawa2009accurate,kanade1994stereo} or neural fields~\cite{mildenhall2021nerf}.
State-of-the-art methods for unordered photo collections use specialized matching, filtering, and bundle adjustment to handle all-pairs matching and dense structure estimation~\cite{schonberger2016structure,snavely2006photo,lou2012matchminer,nister2006scalable,agarwal2011building}. While these approaches have been successful, they remain brittle, as each step can introduce unrecoverable errors.
Recent works have generated 3D models ``zero shot'' solely from models trained solely on 2D images~\cite{chan2022efficient, jain2022zero,poole2023dreamfusion,wang2023score,sanghi2022clip}. \citet{poole2023dreamfusion} and \citet{wang2023score} used the score function of a text-to-image generation model, an approach that they called {\em score distillation sampling}~\cite{poole2023dreamfusion} or {\em score Jacobian chaining}~\cite{wang2023score}. Other work extends this approach with 3D synthetic data for fine-tuning~\cite{shi2023mvdream,liu2023zero} and improved optimization~\cite{lin2023magic3d,wang2023prolificdreamer}. However, these models still face issues like the Janus problem due to difficulties in assigning pose~\cite{poole2023dreamfusion}.
We extend this framework to generate NeRFs that are assigned probability under a GPS-to-image model: we seek a NeRF for which every viewpoint has a high probability under the conditional distribution, avoiding the need for explicit camera pose estimation or feature matching. 

\paragraph{Compositional generation}
A notable characteristic of diffusion models is the ease with which they allow concept composition through the simple addition of noise estimates. This can be interpreted by treating these noise estimates as gradients of a conditional data distribution~\cite{song2019generative,song2020score}, where their sum points in a direction that jointly enhances multiple conditional likelihoods. This technique has been used to enable compositions of text prompts both globally~\cite{liu2022compositional} and spatially~\cite{bar2023multidiffusion,du2023reduce}, of various image transformations~\cite{geng2024visual} and individual image components~\cite{geng2025factorized}. It has also been used for diffusion models originating from two distinct modalities (sight and sound)~\cite{chen2024images}. Another line of work like ControlNet~\cite{zhang2023adding} shows the composition of multiple conditions (\eg, text prompt and pose\&depth). In this paper, we demonstrate that: 1) two conditions of GPS tags and text prompts can successfully generate images using a single noise estimate; 2) our GPS-to-image diffusion models can obtain images representative of a given concept over a large geographic area by averaging noise estimates.

\section{Method}
\label{sec:method}
We propose GPS-to-image diffusion model to synthesize a tourist image by conditioning on GPS coordinates and text. We then show that a variation of the model can be trained on a single tourist site and can be used to ``lift'' 3D models.

\newcommand{\x}[0]{{\mathbf x}}
\begin{figure*}[!t]
\vspace{-2mm}
    \centering
     \includegraphics[width=\linewidth]{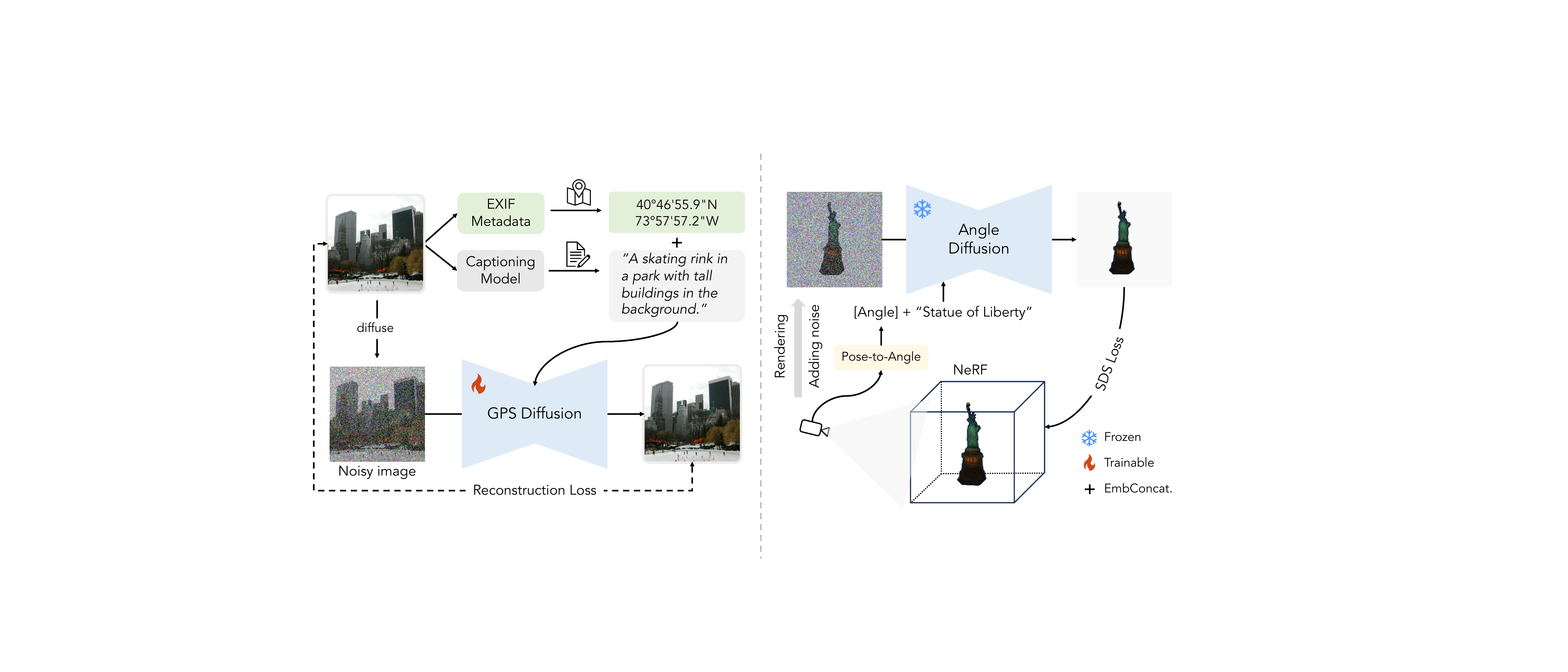}
        \begin{flushleft}
    \small
        \vspace{-3mm}
        \hspace{28mm} (a) GPS-to-image generation \hspace{45mm} (b) GPS-to-3D reconstruction
         \vspace{-2mm}
    \end{flushleft} 
    \caption{{\bf Method.} (a) After downloading geotagged photos, we train a GPS-to-image generation model conditioned on GPS tags and text prompts. The trained generative model can produce images using both conditioning signals in a compositional manner. (b) We can also extract 3D models from a landmark-specific GPS-to-image model using score distillation sampling. This diffusion model parameterizes the GPS location by the azimuth with respect to a given landmark's center. \textbf{+} means we concatenate GPS embeddings and text embeddings.}

    \label{fig:method}
    \vspace{-6mm}
\end{figure*}

\subsection{GPS-to-image diffusion}
\label{gps_diffusion}
We train a model to generate images conditioned on GPS coordinates from a given city, a challenging case that requires capturing fine-grained distinctions between the appearance of different locations. This model is further conditioned on text prompts to improve control of the model.
\paragraph{Preliminaries.} Diffusion models~\cite{dhariwal2021diffusion,ho2020denoising,sohl2015deep,song2020denoising,song2019generative,song2020score} iteratively denoise the Gaussian noise $\mathbf{x}_{T}$ to generate the image $\mathbf{x}_{0}$ of a distribution, which is a reverse of forward process. In the forward process, a clean image $\mathbf{x}_{0}$ is gradually transitioned into random noise $\mathbf{x}_{T}$ by adding Gaussian noise. At each time step, the noisy latent $\mathbf{z}_{t+1}$ can be expressed:
\begin{equation*}
    \mathbf{z}_{t+1} = \alpha_{t}\mathbf{z}_{t} + \beta_{t}\boldsymbol{\epsilon}_{t}\text{,}
\end{equation*}where $\mathbf{z}_{t}$ is noisy latent of previous timestep and $\boldsymbol{\epsilon}_{t}$ is a standard Gaussian noise. $\alpha_{t}$ and $\beta_{t}$ are predefined coefficients, so $ \mathbf{z}_{t+1}$ is also the function of $\mathbf{x}_{0}$.
In DDPM~\cite{ho2020denoising}, the training objective of diffusion models is simplified to:
\begin{equation}
    \mathcal{L}\left(\phi\right) = \mathbb{E}_{t, \mathbf{x}_{0}, \boldsymbol{\epsilon}_{t}}\left[\omega\left(t\right)||\boldsymbol{\epsilon}_{t} - \boldsymbol{\epsilon}_{\phi}\left(\mathbf{z}_{t}, y, t\right) ||^{2}_{2} \right]\text{,}
\end{equation} where $\omega\left(t\right)$ is a weighting function of timestep t (usually set to 1), $\boldsymbol{\epsilon}_{\phi}$ is the denoiser, and $y$ is the condition such as text prompt. For inference, DDIM~\cite{song2020denoising} and classifier-free guidance (CFG) are usually employed:
\begin{equation}
\hat{\boldsymbol{\epsilon}}_{\phi}\left(\mathbf{z}_{t}; y, t\right) = (1 + \omega)\boldsymbol{\epsilon}_{\phi}\left(\mathbf{z}_{t};y, t\right) - \omega\boldsymbol{\epsilon}_{\phi}\left(\mathbf{z}_{t};\varnothing, t\right) \text{,}
\end{equation} 
where $\omega$ is the guidance weight and $\hat{\boldsymbol{\epsilon}}_{\phi}\left(\mathbf{z}_{t}; y, t\right)$ is the predicted noise for denoising.

\paragraph{Training a GPS-conditioned diffusion}
Given a collection of geotagged tourist photos over a certain area (like a city), we want to learn a diffusion model~\cite{rombach2022high,sohl2015deep} that can synthesize tourist images controllably, when conditioned on the photo's GPS position. For a randomly taken tourist image $\x$, we use the GPS coordinates $(x, y)$ to represent its position, where $x$ and $y$ are the longitude and latitude respectively. We use the $(x, y)$ as an extra condition for diffusion models to make them aware of position geospatially. For instance, tourist photos taken at the Louvre Museum usually contain the Louvre Pyramid but not the Arc de Triomphe. In this way, we can endow models with the capability to provide reliable tour guides for walking through Paris by controllably synthesizing tourist photos. 

We build our model on top of a pretrained text-to-image latent diffusion model~\cite{rombach2022high}. Since our off-the-shelf model accepts a text prompt, we provide the text caption generated by BLIP-3~\cite{xue2024xgen} on our collected datasets,  encoded as a CLIP~\cite{radford2021learning} text embedding $\mathbf{p} \in \mathbb{R}^{L \times D}$, where $L$ is the number of text tokens and $D$ is the token feature dimension.
We then learn a GPS-pose embedding $\mathbf{g} = \left[f(x), f(y)\right] \in \mathbb{R}^{2 \times D}$, and append it to text tokens $\mathbf{p}$ to establish a ``GPS" CLIP text condition, as shown in \cref{fig:method}. This input representation ensures that the model starts from an initialization that closely resembles what it was trained on.

We finetune the model using the text embedding $\mathbf{p}$ and GPS embedding $\mathbf{g}$. Specifically, given a tourist photo dataset $\mathcal{X}$, for training samples $\{\mathbf{x}, \mathbf{p}, \mathbf{g}\}$, we optimize the diffusion loss:
\begin{equation}\label{recon_loss}
    \mathcal{L}_{\mathrm{recon}} = \mathbb{E}_{\mathbf{x}, \mathbf{g}, \boldsymbol{\epsilon}_{t}, t} \left[\parallel \boldsymbol{\epsilon}_{t} - \boldsymbol{\epsilon}_{\phi}(\mathbf{z}_{t}; \mathbf{p}, \mathbf{g}, t) \parallel_{2}^{2}\right] \text{,}
\end{equation}
where $\mathbf{z}_t$ is the noisy latent of image $\x$ at timestep $t$. 

\paragraph{Inference}During inference, we use the classifier-free guidance strategy from InstructPix2Pix~\cite{brooks2023instructpix2pix} for two conditions~(text prompt and GPS tag). Our score estimate is as follows:
\begin{align}\label{dual_inference}
    \tilde{\boldsymbol{\epsilon}}_{\phi}\left(\mathbf{z}_{t}; \mathbf{p}, \mathbf{g}, t\right) = &\boldsymbol{\epsilon}_{\phi}\left(\mathbf{z}_{t};\varnothing, \varnothing, t\right) \notag \\
    &+ \omega_{\mathbf{p}}\left(\boldsymbol{\epsilon}_{\phi}\left(\mathbf{z}_{t};\mathbf{p}, \varnothing, t\right) - \boldsymbol{\epsilon}_{\phi}\left(\mathbf{z}_{t};\varnothing, \varnothing, t\right)\right) \notag \\
    &+ \omega_{\mathbf{g}}\left(\boldsymbol{\epsilon}_{\phi}\left(\mathbf{z}_{t};\mathbf{p},\mathbf{g}, t\right) - \boldsymbol{\epsilon}_{\phi}\left(\mathbf{z}_{t};\mathbf{p}, \varnothing, t\right)\right)\text{,} 
\end{align}
where $\omega_{\mathbf{p}}$ and $\omega_{\mathbf{g}}$ are gudiance weights.

\subsection{GPS-guided 3D reconstruction}\label{subsec:3d}
Recent work has shown that 3D models can be extracted from 2D text-to-image diffusion models. We build on this idea to obtain 3D reconstructions of specific locations using GPS-to-image models.
\paragraph{Preliminary} Prior work~\cite{poole2023dreamfusion,wang2023score,wang2023prolificdreamer,shi2023mvdream,lin2023magic3d} leverages pretrained 2D text-to-image diffusion models like Imagen~\cite{saharia2022photorealistic} to synthesize 3D contents by textual descriptions. During optimization, a Gaussian noise would be added to each NeRF rendering $\mathbf{x} = h_{\theta}\left(\mathbf{q}\right)$, where $\mathbf{q}$ is the camera pose. Then noisy rendering would be fed into pretrained denoiser $\boldsymbol{\epsilon}_{\phi}$ and score distillation sampling (SDS) loss provides gradient to guide NeRF~\cite{mildenhall2021nerf,barron2022mipnerf360} training:
{\small
\begin{equation}
    \nabla_{\mathbf{\theta}}\mathcal{L}_{SDS}\left(\phi, \mathbf{x}=h_\theta\left(\mathbf{q}\right)\right) \approx 
    \mathbb{E}_{t, \epsilon} \left[\omega\left(t\right)\left(\hat{\epsilon}_{\phi}\left(\mathbf{z}_{t}; y, t\right) - \epsilon\right)\frac{\partial \mathbf{x}}{\partial \mathbf{\theta}}\right] \text{.}
\end{equation}}
The text prompt $y$ is appended by view-dependent texts of ``front view'', ``side view'', or ``back view'' based on the randomly sampled camera poses $\mathbf{q}$, which can benefit 3D generation results. However, in some cases, 2D text-to-image diffusion models struggle to control viewpoints accurately, causing a multi-face Janus issue~\cite{poole2023dreamfusion}.

\paragraph{Angle-to-image diffusion}
GPS signals provide useful information, \eg, viewpoint details, for 3D landmark reconstruction from tourism photos. Hence, we train a diffusion model to transform these implicit signals into a score function for supervision.
For an image $\x$ taken from a landmark at GPS coordinate $(x, y) \in \mathbb{R}^2$, we parameterize the pose using the azimuth angle $\alpha$, with respect to the center of the landmark $(x_{o}, y_{o})$:    $\alpha = \arctan \left(\frac{x - x_{o}}{y - y_{o}}\right)$. Here we use $\alpha$ instead of $ (x, y)$ as an extra condition for diffusion models, which means that we would replace $\mathbf{g}$ in~\cref{recon_loss} with $\mathbf{g\prime}=f'(\alpha)\in\mathbb{R}^{1\times D}$.
This representation makes it straightforward to combine the approach with DreamFusion~\cite{poole2023dreamfusion}, which can be easily extended to accept angular conditioning. Additionally, we fix the text prompt to ``{\tt A photo of \{landmark name}\}" for each landmark. We refer to the diffusion model conditioned on GPS coordinates as the GPS-to-image diffusion, and the model conditioned on the angle from GPS tags as the angle-to-image diffusion.

We found that when the focused area is small, by directly finetuning the diffusion model with only a few thousand self-collected tourist photos, the model adapts to the distribution very easily and sometimes loses the generative diversity of the original model.  To avoid this issue, we follow \cite{ruiz2023dreambooth} and use a {\em prior preservation} loss to maintain the prior knowledge and regularize the pose-conditioned training. More details are presented in Appendix~\ref{subsec:angle_supp}. We combine both losses and optimize the final objective:
\begin{equation}\label{total_loss}
    \mathcal{L} = \mathcal{L}_{\mathrm{recon}} + \lambda  \mathcal{L}_{\mathrm{preservation}} \text{,}
\end{equation}
where $\lambda$ is the weight to balance the reconstruction and preservation loss.

\begin{figure}[t]
\centering
    
  \includegraphics[width=\linewidth]{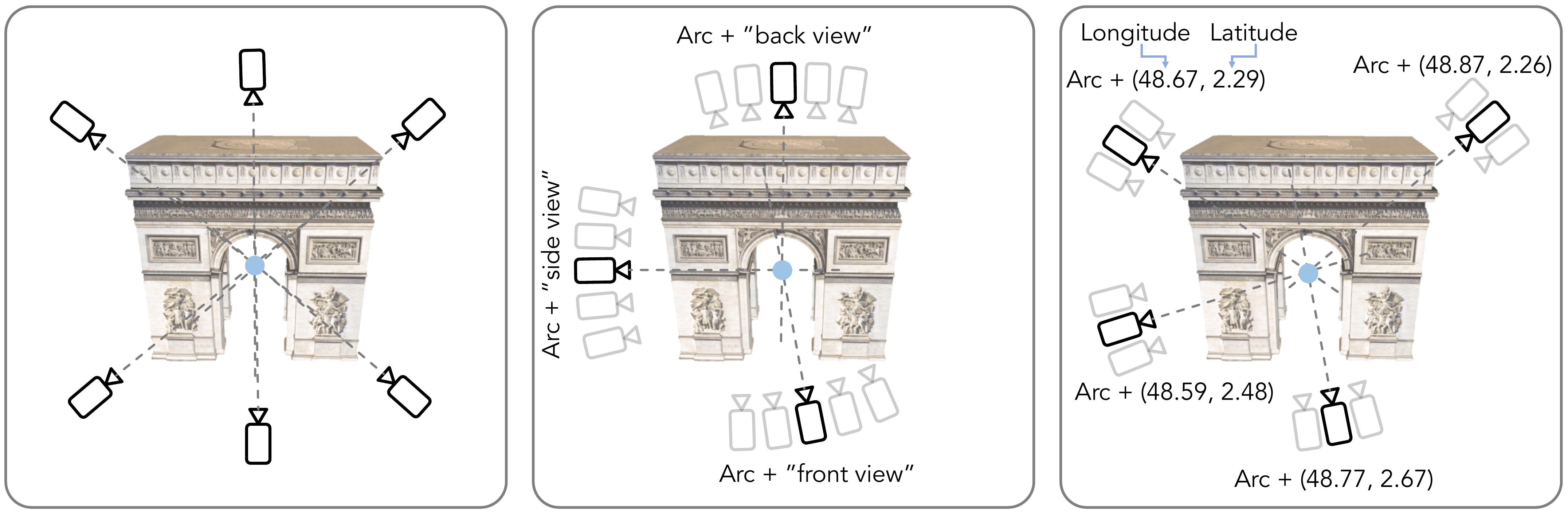}
    
    \begin{flushleft}
    \small
        \vspace{-4.5mm}
        \hspace{8mm} (a) SfM \hspace{10mm} (b) DreamFusion \hspace{10mm} (c) Ours

         \vspace{-2.75mm}
    \end{flushleft} 
        \caption{{\bf 3D Setup Comparison.} We extract 3D models from 2D GPS-to-image models. (a) Traditional approaches require running SfM to estimate camera pose, followed by dense geometry estimation. Since they are based on triangulation, they are susceptible to catastrophic errors due to incorrect pose; (b) DreamFusion~\cite{poole2023dreamfusion} samples images from different poses within a scene using view-dependent prompting. However, text has a limited ability to precisely control the position of the camera. (c) Our method extends DreamFusion with GPS conditioning,  reducing pose uncertainty.}
    \vspace{-5mm}

    \label{fig:high-level}
\end{figure}

\paragraph{GPS-guided score distillation sampling}
Using the pose information from GPS, our angle-to-image diffusion model generates photos of monuments from various viewpoints by conditioning on the GPS poses. Our angle conditioning is analogous to view-specific prompting in Poole \etal~\cite{poole2023dreamfusion} while providing a better view prior, as shown in \cref{fig:high-level}.
We use score distillation sampling (SDS) to extract a 3D model of the landmark from our diffusion model~\cite{poole2023dreamfusion} (\cref{fig:method}(b)). We parameterize this 3D model as a NeRF $h_\theta\left(\cdot\right)$ with parameters $\theta$. We optimize the parameters of a NeRF~\cite{mildenhall2021nerf} using gradient descent, such that every rendered viewpoint has a high probability under the Angle-conditioned image model. 

During each iteration of the optimization process, we sample random virtual camera poses. For each one, we transform the pose $\mathbf{q}$ to azimuth $\alpha$, and create a conditioning embedding $\mathbf{g\prime}$ (\cref{fig:method} (b)). We can then render a corresponding image $\mathbf{x} = h_{\theta}\left(\mathbf{q}\right)$ from the NeRF.

We use the SDS loss to obtain a gradient from the angle-to-image diffusion model to supervise NeRF~\cite{muller2022instant}. Following recent approaches~\cite{shi2023mvdream,huang2023dreamtime}, we gradually decrease both the maximum and minimum of the time step sampling interval for the SDS loss during the optimization process. We apply classifier-free guidance (CFG)~\cite{ho2022classifier} to the SDS loss. For the unconditional version of the model, we simultaneously zeroed out both text and GPS conditions. This results in a noise estimator:
\begin{equation}\label{cfg}
    \hat{\boldsymbol{\epsilon}}_{\phi}\left(\mathbf{z}_{t}; \mathbf{p}, \mathbf{g\prime}, t\right) = (1 + \omega)\boldsymbol{\epsilon}_{\phi}\left(\mathbf{z}_{t};\mathbf{p}, \mathbf{g\prime}, t\right) - \omega\boldsymbol{\epsilon}_{\phi}\left(\mathbf{z}_{t};\varnothing, \varnothing, t\right) \text{,}
\end{equation}
where $\mathbf{z}_{t}$ is the noisy latent of image rendered by NeRF and $\omega$ is the CFG guidance weight. The equation of gradient is presented in Appendix~\ref{3d_supp}.   

\begin{figure}[b]
    \centering
     \includegraphics[width=\linewidth]{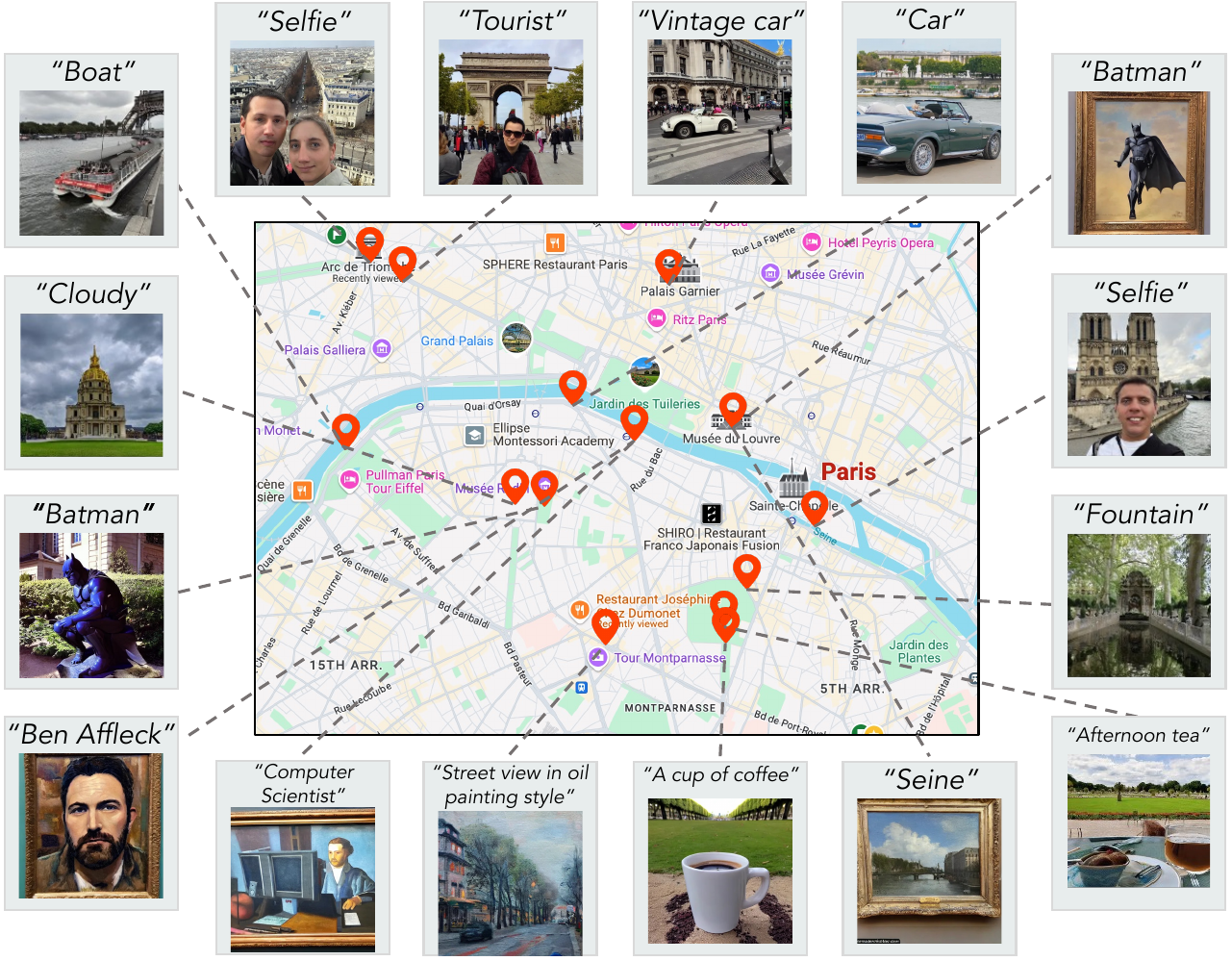}
   \caption{{\bf Qualitative results for Paris.} We show images that have been sampled from our GPS-to-image diffusion model for various locations and prompts within Paris.}
    \label{fig:paris_map}
    \vspace{-5mm}
\end{figure}

\section{Experiments}
\label{sec:experiments}
\begin{figure*}[!t]
\vspace{-2mm}
    \centering
     \includegraphics[width=\linewidth]{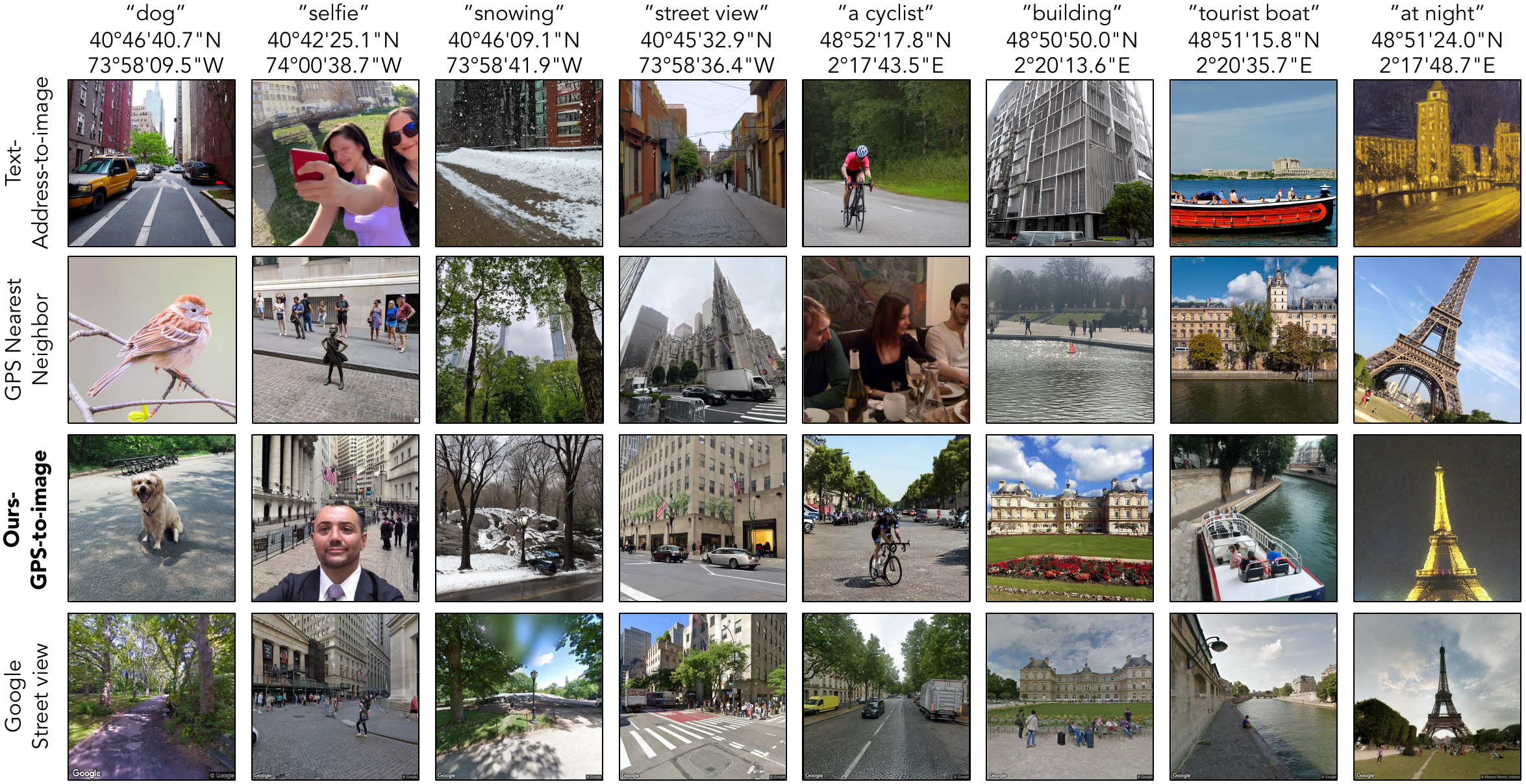}
    \caption{{\bf Qualitative comparison for GPS-to-image diffusion.} We compare the qualitative results of our method against baselines using specific pairs of text prompts and GPS tags. Each column shows a text prompt and a GPS tag at the top. Text-address-to-image diffusion model is conditioned on a combination of the text prompt and the address name derived from the GPS tag. We also perform nearest neighbor in the training set based on GPS tags. Our GPS-to-image diffusion model uses a text prompt and raw GPS tag as conditioning. Google Street View images are sampled for \textbf{reference} of geolocation. Our method achieves better compositionality and visual quality.}

    \label{fig:diff_qualitative}
    \vspace{-3mm}
\end{figure*}

We evaluate our model using a variety of quantitative and qualitative metrics. We first evaluate our GPS-to-image diffusion model, measuring its ability to successfully generate images that convey GPS and semantics of text prompts. We then evaluate our model's ability to obtain 3D reconstructions for landmarks guided by GPS.

\subsection{Implementation details}\label{photo_collection}

\paragraph{Tourist photo collection} 
To train GPS-to-image diffusion models, we obtained two city photo collections with GPS tags by querying from Flickr: \textbf{1)} New York City~(Manhattan, 501,592 photos); \textbf{2)} Paris~(315,306 photos). For the landmark reconstruction task, we gather 6 sets of landmark photos following a similar approach. The number of evaluated landmarks aligns with previous work in the field~\cite{martinbrualla2020nerfw}. Please see Appendix~\ref{supp:dataset} for dataset details.

\paragraph{GPS-to-image diffusion.} We use a positional encoding with frequencies of 10~\cite{mildenhall2021nerf} and a two-layer MLP to encode the GPS conditions. For each city, we normalize $(x,y)$ to the range $[-1, 1]$. We finetune Stable Diffusion-v1.4~\cite{rombach2022high} on Flickr images, at a resolution of $512\times512$ for 15k steps. We use the AdamW~\cite{loshchilov2017decoupled} optimizer with a learning rate of $1\times10^{-4}$. We use a batch size of 512 on 8 NVIDIA L40S GPUs. During training, we randomly drop text and GPS conditions, ensuring 5\% text-only, 5\% GPS-only, and 5\% unconditional generation. We caption images using BLIP-3~\cite{xue2024xgen}. See Appendix~\ref{supp:gps-to-image-details} for details.

\paragraph{Angle-to-image diffusion.} 
We train individual Angle-to-diffusion models for each landmark. We calculate the azimuth angle $\alpha$ from GPS using $\alpha = \arctan \left(\frac{x - x_{o}}{y - y_{o}}\right)$ and map it to the nearest 10° angle bin. We use the normalized bin value with positional encoding as conditional input.
We fix text prompts with the template ``{\tt A photo of \{landmark name\}}''. The weight of preservation loss $\lambda$ in \cref{total_loss} is set to 1.0. Please refer to Appendix~\ref{supp:angle-to-image details} for more details.

\paragraph{3D reconstruction} 
We apply different guidance weights of classifier-free guidance (CFG) in \cref{cfg} for score distillation sampling for each landmark.
We turn on shading after 1000 steps and use orientation and opacity regularization loss, following \cite{poole2023dreamfusion}. We use the Adam optimizer~\cite{kingma2014adam} with a learning rate of 0.01 for 10k training steps. The time step sampling interval is gradually reduced from $[0.98, 0.98]$ to $[0.02, 0.50]$. 
To match the characteristics of tourism photo distribution, we restrict the elevation angle of sampled virtual camera views to below 0, while the azimuth angle is sampled across the full range.

\subsection{Evaluation of GPS-to-image generation}\label{subsec:gps2img}

\begin{table}[!t]
\newcommand\padd{\phantom{0}}
\centering
\small
\caption{{\bf Evaluation of GPS-to-image diffusion.} We compare our method with several baselines in terms of CLIP Score and GPS Score. NN represents the nearest neighbor and SD is for stable diffusion. The best results are in \textbf{bold}, and the second bests are colored in blue.}
\resizebox{\linewidth}{!}{
\begin{tabular}{lccc}
\toprule
\textbf{Method}  &  CLIP Score~($\uparrow$)  &  GPS Score~($\uparrow$)  &  Avg ($\uparrow$)\\
\toprule
GPS NN &    18.77    & \textcolor{blue}{13.66} &  \textcolor{blue}{16.22} \\
SD (Text\&address)
  &    26.65   & \padd 4.25  & 15.45 \\
SD (Text) 
  &   \textbf{29.13}    &  \padd 1.21  & 15.17\\
\cdashlinelr{1-4}
Ours &  \textcolor{blue}{27.88} & \padd 8.15  & \textbf{18.02} \\ 
Ours~(w/o text) & --  &  \textbf{13.71}  & -- \\ 
\bottomrule
\end{tabular} }
\label{tab:gps2img_quantitative}
\vspace{-4mm}
\end{table}

\begin{figure*}[t]
    \centering
     \includegraphics[width=\linewidth]{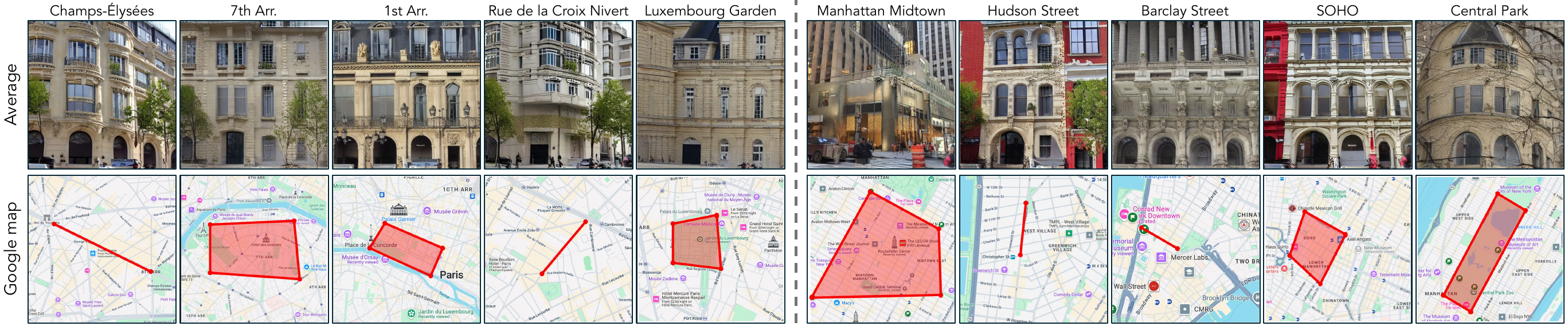}
        \begin{flushleft}
    \small
        \vspace{-3mm}
        \hspace{40mm} (a) Paris \hspace{68mm} (b) New York City
         \vspace{-3mm}
    \end{flushleft} 
    \caption{{\bf Average images.} We select five areas for Paris and New York City respectively. Using our GPS-to-image models, we obtain representative images of the concept of ``{\tt building}'' within these geographic regions to observe architectural styles. More examples can be found on \href{https://cfeng16.github.io/gps-gen/}{project webpage} for different locations and concepts.}
    \label{fig:average}
\end{figure*}

We first evaluate our GPS-to-image diffusion models in generating photos conditioned on GPS tags.

\paragraph{Evaluation metrics.} 
To evaluate our model and baselines, we create 1,292 random GPS-text pairs as conditions for generation. We report CLIP score (CS)~\cite{radford2021learning} to measure the alignment between generated image and text prompts. Analogously, we train a GPS-CLIP model on paired GPS-image data with contrastive loss ~\cite{chen2020simple,radford2021learning,he2020momentum,oord2018representation} and report GPS score~(GS) which measures cosine similarity between image and GPS embeddings. See Appendix~\ref{supp:gps-clip} for details.

\paragraph{Baselines.} Since we are not aware of any prior work on GPS-to-image generation, we include two baselines: \textbf{1)} Stable Diffusion (SD)v1.4~\cite{rombach2022high}; \textbf{2)} GPS Nearest Neighbor. For SD, we consider two variations: the first accepts a concatenation of the text prompt and address name\footnote{The address name is obtained through geocoding using \href{https://geopy.readthedocs.io/en/stable/}{GeoPy}.}, while the second is conditioned only on the text prompt.

\paragraph{Results}
As shown in \cref{tab:gps2img_quantitative} and \cref{fig:diff_qualitative}, our method achieves the best performance in terms of the average CLIP score and GPS score. Additionally, it demonstrates better compositionality, indicating that our method can successfully generate images from text prompts and GPS tags. Our method, without the text prompt, achieves a better GPS score than the Nearest Neighbor, indicating our model better captures image distributions in a geospatial context. More qualitative results are presented in~\cref{fig:teaser} and~\cref{fig:paris_map}. For instance, our method can successfully generate images under different weather~\cite{li2023climatenerf} and lighting~\cite{liu2020learning} conditions given a certain GPS location.

\subsection{Average images}
Inspired by work that computes average images~\cite{doersch2012makes,torralba2003statistics}, we apply our GPS-to-image models to the problem of obtaining images that are representative of a given concept over a large geographic area. Specifically, we generate a single image that has high probability under all GPS locations within a user-specified area, as measured by our diffusion model. To do this, we follow work on compositional generation~\cite{liu2022compositional} and simultaneously estimate noise using a large number of GPS locations and average together the noise vectors during each step of the reverse diffusion process (see Appendix~\ref{supp:avg} for details). In~\cref{fig:average}, we show images generated for the text prompt ``{\tt building}'' over a variety of streets in neighborhoods in Paris and New York. The resulting images capture the distinctive architectural styles of buildings in the specified areas. More examples can be found on the \href{https://cfeng16.github.io/gps-gen/}{project webpage}, showcasing different locations and concepts.

\begin{figure*}[t]
    \centering
    \vspace{-2mm}
     \includegraphics[width=\linewidth]{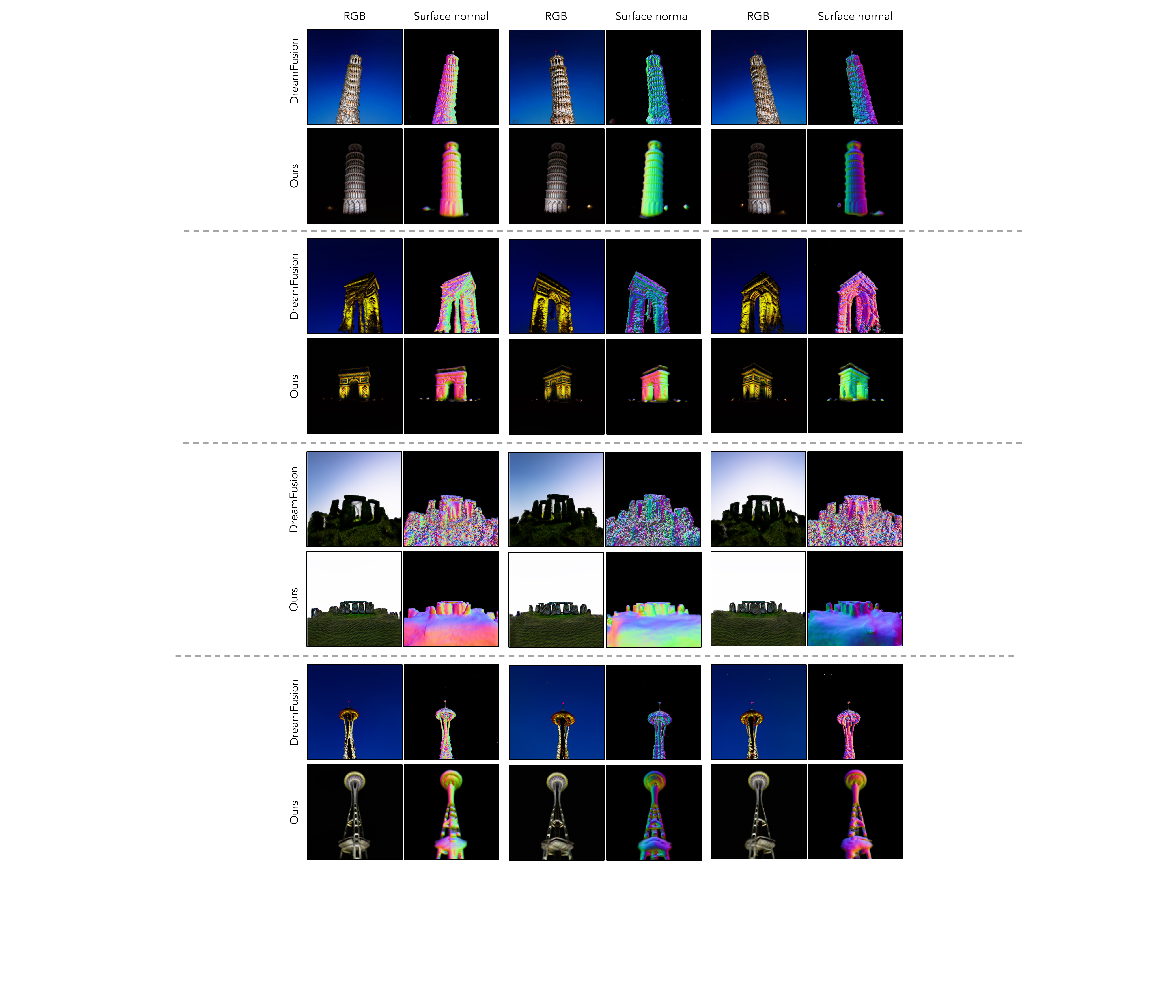}
    \caption{{\bf Qualitative comparison for 3D monument reconstruction.} We show qualitative results of DreamFusion~\cite{poole2023dreamfusion} and our method on two monuments: 1) Leaning Tower of Pisa; 2) Arc de Triomphe. Our reconstructed 3D monuments have better visual quality and more accurate 3D structure. We use rendered depth to make the background of RGB rendering white. {\bf Please see Appendix~\ref{supp:gps-to-3d results} and \href{https://cfeng16.github.io/gps-gen/}{project webpage} for more examples.}} 
    \label{fig:3d_recon}
    \vspace{-2.5mm}
\end{figure*}

\subsection{Evaluation of angle-to-image generation}

We also evaluate how well our angle-to-image diffusion models can generate photos of monuments conditioned on desired viewpoints~(angles) derived from GPS tags.

\paragraph{Evaluation metric} 
We train a classifier on each landmark dataset individually to predict the discretized angle bins derived from GPS tags. For each angle bin of 10°, we ask generative models to synthesize 10 images and pair them up with input angle bins as ground truth. Then we apply the trained angle classifier to evaluate these images using accuracy as the metric. The evaluation method for diffusion models we adopt is similar to CLIP score~\cite{radford2021learning}. We use this classifier trained on our training dataset to testify whether the finetuned diffusion model has successfully fit the training distribution. It should be noted that the main goal of finetuning a diffusion model is to facilitate the end goal of 3D landmark reconstruction---the reconstruction quality of the reconstructed 3D landmarks should serve as an evaluation that our model has successfully learned the data distribution and angle conditioning signal from GPS.

\begin{table}[!t]
\centering
\small
\vspace{-1mm}
\caption{{\bf Evaluation of angle-to-image diffusion. } We evaluate the accuracy of our model in generating images with the correct azimuth, as determined by an image-to-azimuth model.} 
\begin{tabularx}{0.85\linewidth}{Xc}
\toprule
\textbf{Method}  & Angle acc ($\%$) \\
\toprule
Random chance              &  2.78  \\
 Stable Diffusion~\cite{rombach2022high}   &    3.06        \\
Ours &   \textbf{22.36}       \\
\bottomrule
\end{tabularx}
\label{tab:angle_acc}
\vspace{-2mm}
\end{table}

\begin{table}[!t]
\centering
\small
\caption{{\bf  Quantitative comparison for 3D.} We report results: CLIP Score (CS)~\cite{radford2021learning}, Perceptual Quality (PC), and Tourist Score (TS). It shows that our method achieves the highest quality.} 
\begin{tabular}{lccc }
\toprule

\textbf{Method}  & CS~\cite{radford2021learning} ($\uparrow$)  &  PQ ($\uparrow$)& TS ($\uparrow$)\\
\toprule
NeRF~\cite{tancik2023nerfstudio}   &  20.57 &  1.32 & 1.36\\
 Dreamfusion~\cite{poole2023dreamfusion}   &    29.49     &  2.21  & 2.09  \\

Ours & \textbf{31.87}  & \textbf{3.31}  & \textbf{3.45}  \\ 
\bottomrule
\end{tabular}
\label{tab:3d_human_study}
\vspace{-4mm}
\end{table}

\paragraph{Results}
We compare our method with two baselines: \textbf{1)} Stable Diffusion (SD) v1.4~\cite{rombach2022high}; \textbf{2)} Random chance. The results are reported in   \cref{tab:angle_acc}. 
Our angle-to-image diffusion model significantly outperforms both random chance and text-to-image Stable Diffusion~\cite{rombach2022high} by a relatively large margin, demonstrating that it effectively learns viewpoint signals from image-GPS pairs. Some generated images from the model are provided in Appendix~\ref{supp:angle-to-image results}.

\subsection{Evaluation of 3D landmark reconstruction}
We evaluate how good our reconstructed landmarks are based on our angle-to-image diffusion model. 
\paragraph{Evaluation metrics.} The generated 3D landmarks are evaluated both qualitatively and quantitatively. We calculate CLIP Score (CS)~\cite{radford2021learning} on RGB renderings of the landmarks with their corresponding text prompts of their names. The final CLIP score is presented as an average calculated from 30 randomly selected views. Following prior work~\cite{hollein2023text}, we also conducted a user study for evaluation. We asked 36 participants to score the Perceptual Quality (PQ) and Tourist Score (TS) of landmark reconstructions on a scale of 1 to 5, where 5 is the best.
We define the perceptual quality metric to evaluate the quality of generated 3D assets, which do not necessarily match ground truth. The tourist score evaluates the 3D landmark reconstruction compared with real tourist photos by human preference. We evaluate 6 landmarks as mentioned in \cref{photo_collection}.

\paragraph{Baselines.} We compare our method to DreamFusion~\cite{poole2023dreamfusion} and COLMAP~\cite{schoenberger2016sfm} followed by NeRF~\cite{mildenhall2021nerf}. For the latter step, we train Nerf in the wild (NeRF-W)~\cite{martin2021nerf}~\footnote{We use this popular \href{https://github.com/kwea123/nerf_pl/tree/nerfw}{reimplementation} since NeRF-W is not publicly available.} and Nerfacto~\cite{tancik2023nerfstudio} for each scene. For DreamFusion~\cite{poole2023dreamfusion}, we use Stable Diffusion~\cite{rombach2022high} to ensure a fair comparison with our model. We consider two types of text prompts: 1) ``{\tt \small A photo of \{landmark name\}}"; 2) ``{\tt \small\{landmark name\}}", both of them are with view-dependent conditioning (text prompt is appended with ``{\tt \small front/back/side view}"). We pick the best one for evaluation.

\paragraph{Results}
As shown in \cref{tab:3d_human_study}, our method outperforms two baselines in terms of three evaluation metrics.  
Our qualitative results in \cref{fig:3d_recon} show that renderings from our models have better visual quality and more accurate 3D structure than those from DreamFusion~\cite{poole2023dreamfusion}. This suggests that our GPS-conditioned diffusion model can provide a better pose prior than the text-to-image diffusion~\cite{rombach2022high} with view-related prompts.
As expected, we found the SfM-based baseline to be ``all or nothing'', either providing very high-quality reconstructions or catastrophically failing (as shown in Appendix~\ref{subsec:nerf-sfm}). For example, COLMAP~\cite{schoenberger2016sfm} successfully reconstructs camera poses and sparse point clouds for 3 of the 6 scenes, and fails on three. NeRF-W~\cite{martin2021nerf} estimation completely fails on 6 landmarks and Nerfacto~\cite{tancik2023nerfstudio} fails on 5. This also reveals one shortcoming: if COLMAP~\cite{schoenberger2016sfm} cannot reconstruct the poses of the input images, NeRF~\cite{mildenhall2021nerf} optimization is not possible. In contrast, our method addresses this and is able to reconstruct scenes that COLMAP~\cite{schoenberger2016sfm} cannot reconstruct.

\begin{figure}[!t]
    \centering
     \includegraphics[width=0.83\linewidth]{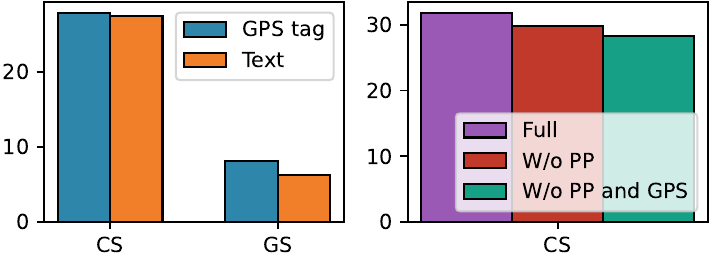}
       \begin{flushleft}
    \small
        \vspace{-3mm}
        \hspace{1mm} (a) Representation of geolocation  \hspace{3mm} (b) 3D reconstruction
         \vspace{-3mm}
    \end{flushleft} 
    \caption{{\bf Ablation.} We conducted ablation studies to analyze the effectiveness of different modules in our method for GPS-to-image generation and 3D landmark reconstruction.} 
    \label{fig:ablation_1}
    \vspace{-4mm}
\end{figure}

\vspace{-2mm}\subsection{Ablation Study} 
\vspace{-2mm}
\paragraph{Attention map visualization}

\begin{figure}[!t]
    \centering
     \includegraphics[width=\linewidth]{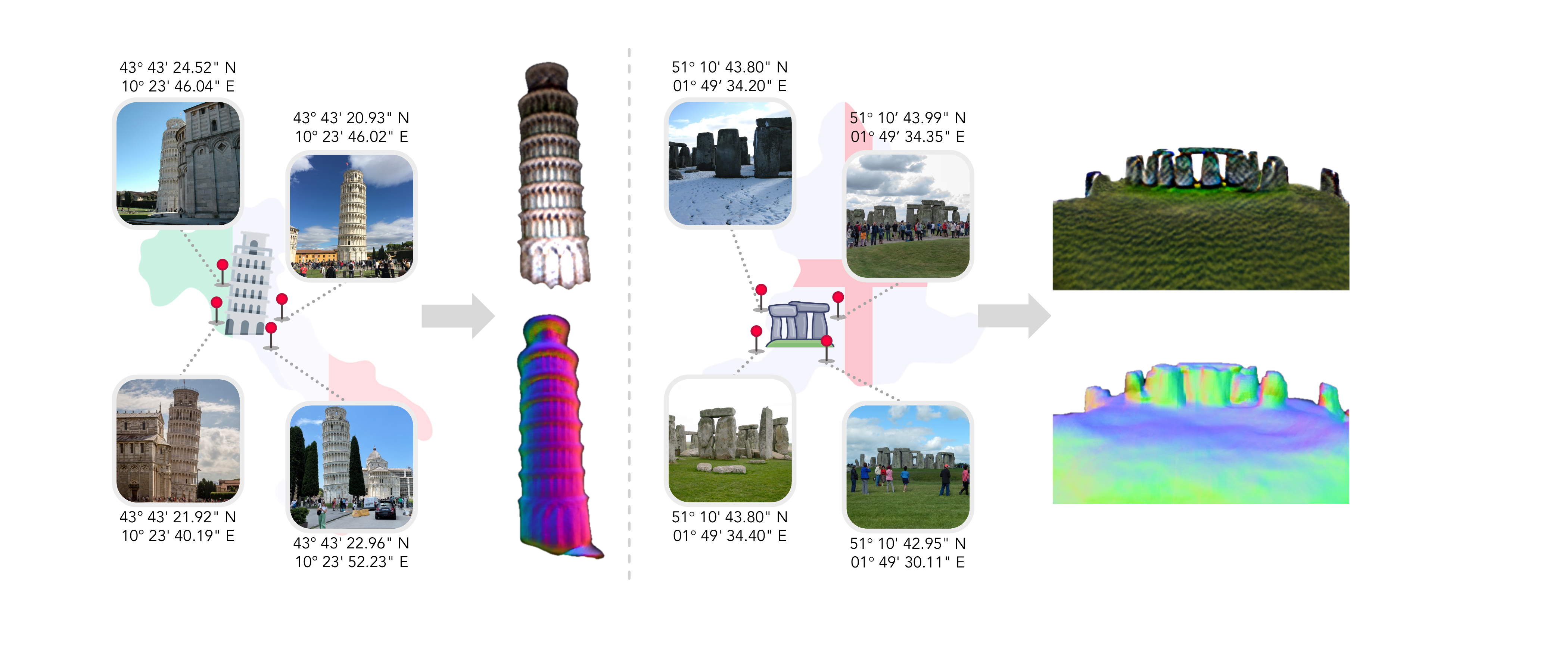}
    \caption{{\bf Attention visualization.} We visualize attention maps for text and GPS tokens.} 
    \label{fig:attention}
\end{figure}

We visualize attention maps~\cite{hertz2022prompt} for the text and GPS conditions to examine what signals the model focuses on. We show two examples in \cref{fig:attention}. We can see that the text prompt effectively controls the semantics of objects in the synthesized image, while the GPS tag (latitude and longitude) significantly influences the background. For instance, in the ``{\tt \small A tourist}" example, we can observe the shape of the Oculus Center in the attention maps corresponding to the GPS tag.

\paragraph{Representation of geolocation.} 
Geolocation can be represented in two variations: 1) continuous GPS tag; 2) address name in text geodecoded from GPS tag. We finetune stable diffusion~\cite{rombach2022high} on these two variations and results are presented in~\cref{fig:ablation_1}. As shown in~\cref{fig:ablation_1}, though CLIP Scores are comparable, our method based on continuous GPS tag outperforms the text-based method by a significant margin for GPS Score. This suggests that using a continuous GPS tag as a conditioning input better controls the geospatial aspects of image generation.
\vspace{-1mm}
\paragraph{3D reconstruction} We conduct experiments to evaluate the importance of prior preservation loss and GPS conditioning for 3D landmark reconstruction. We train our angle-to-image diffusion models without prior preservation loss and also perform experiments where we remove angle conditioning during training (\cref{fig:ablation_1}). Our method outperforms these baselines by a large margin, suggesting that GPS is a valuable conditioning signal for reconstruction.

\section{Conclusion} 
Our work suggests that GPS coordinates are a useful signal for controllable image generation. We have proposed a method to generate images conditioned on GPS tag and text prompt in a compositional manner, which successfully learns the cross-modal association between GPS tags and images. It can achieve compositional generation for tasks that require a fine-grained understanding of how images vary within a city. We also find that GPS conditioning enables us to reconstruct 3D landmarks by score distillation sampling without explicit camera pose estimation. Our work opens two future directions. The first is to develop models that use GPS-to-image generation methods to analyze geotagged photo collections in additional ways. The second is to develop new generative models that can extract more information from GPS conditioning.

\paragraph{Limitations} Our approach may not be well-suited for cases where few photos have GPS available. Score distillation sampling is known to produce saturated images. In some scenarios, GPS tags carry certain semantic information that is difficult to fully disentangle.

\paragraph{Acknowledgements} We thank David Fouhey, David Crandall, Ayush Shrivastava, Chenhao Zheng, Daniel Geng, and Jeongsoo Park for the helpful discussions. We thank Yiming Dou for helping set up NeRF baselines. Chao especially thanks Xinyu Zhang for her help in this project. This work was supported in part by Cisco Systems, NSF CAREER \#2339071, and DARPA Contract No. HR001120C0123.

{
    \small
    \bibliographystyle{ieeenat_fullname}
    \bibliography{main}
}

\clearpage
\renewcommand{\thesection}{A.\arabic{section}}
\setcounter{section}{0}
\begin{strip}
\vspace{-2mm}
    \centering
     \includegraphics[width=\linewidth]{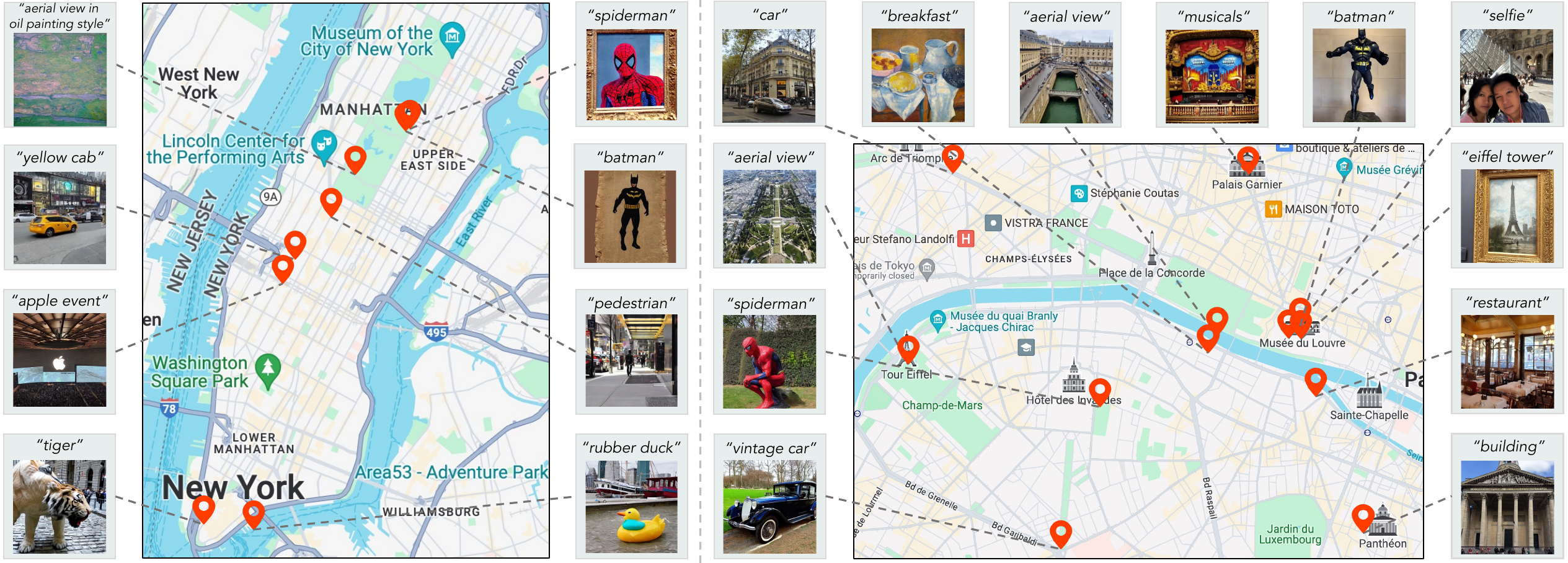}
    \begin{flushleft}
    \small
        \vspace{-3mm}
        \hspace{25mm} (a) New York City \hspace{75mm} (b) Paris
         \vspace{-2mm}
    \end{flushleft} 
    \captionof{figure}{{\bf More qualitative results for GPS-to-image generation.} We present more qualitative results of GPS-to-image generation for New York City and Paris. Images are sampled from a variety of GPS locations and text prompts.}

    \label{fig:diff_qualitative_supp}
\end{strip}

\begin{figure*}[!t]
    \centering
     \includegraphics[width=\linewidth]{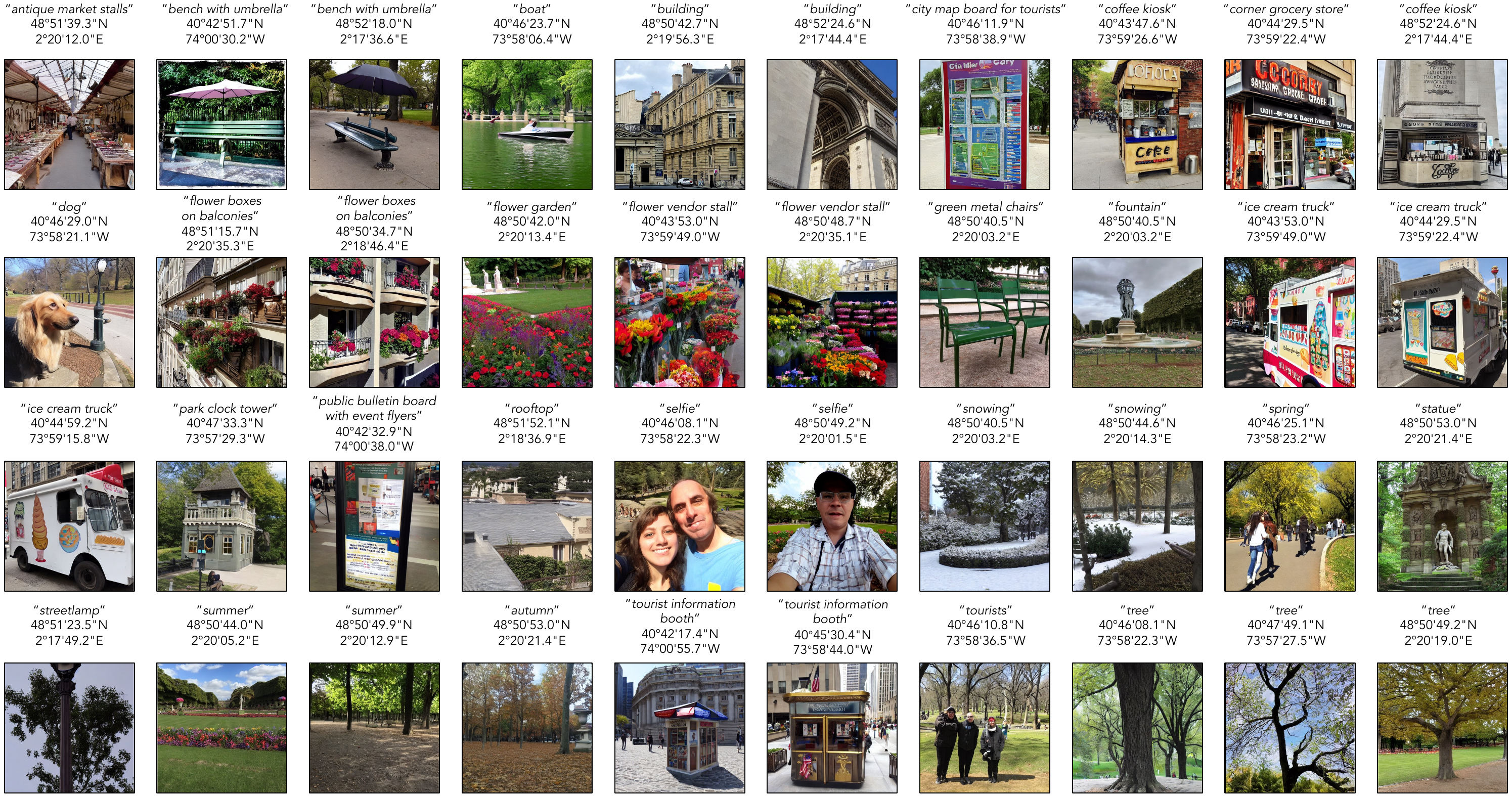}
    \caption{{\bf Random sampling.} We show some randomly sampled images from generation results of our GPS-to-image diffusion models conditioned on text prompts and GPS tags. These sampled results were used in the quantitative evaluation.} 
    \label{fig:random_sample}
\end{figure*}

\section{GPS-to-image generation}
\subsection{More qualitative results}\label{supp:more_qua_results}
We present more qualitative results for our GPS-to-image generation in~\cref{fig:diff_qualitative_supp}. As shown in~\cref{fig:diff_qualitative_supp}, our method can successfully generate images conditioned on GPS tag and text prompt in a compositional manner. For instance, in {\bf New York City~(a)}: 1) text prompt ``{\tt tiger}'' along with GPS location of the Charging Bull statue generates an image of a tiger in a similar pose to the Charging Bull, with an appropriately matching background; 2) given text prompt ``{\tt spiderman}'' and ``{\tt batman}'', we can generate an image of either an oil painting of spiderman or a stele of batman, depending on the location within The Metropolitan Museum of Art; 3) when conditioned on the GPS location of Madison Square Garden and the text prompt ``{\tt apple event}'', our model generates an image that appears to have been taken at Madison Square Garden (see ceiling in the image) during a real Apple event (see Apple logo in the image). In {\bf Paris~(b)}: 1) with text prompt ``{\tt spiderman}'' and GPS location of Rodin Museum, the GPS-to-image diffusion model can generate an image of spiderman statue posed similarly to The Thinker; 2) text prompt ``{\tt batman}'' and GPS location of Louvre Museum's statue gallery can result in an image of statue of batman, while the model can generate an image of painting about Eiffel Tower when GPS location is the painting gallery of Louvre and text prompt is ``{\tt eiffel tower}''; 3) given the text prompt ``{\tt musicals}'' and the GPS location of Palais Garnier, an image depicting a musical performance at Palais Garnier is generated; 4) when conditioned on text prompt ``{\tt breakfast}'' and GPS location of Orsay Museum, our GPS-to-image diffusion model can generate an image of oil painting of breakfast; 5) using the text prompt ``{\tt car}'' along with the GPS location of the Champs-Élysées, an image of the car is generated with a background filled with buildings in the Haussmannian architectural style. Additionally, we present randomly sampled images from generation results of our GPS-to-image diffusion models in~\cref{fig:random_sample}. Some images have visible artifacts, and this may be due to the limited availability of photos with GPS tags in that area.

\subsection{Evaluation set}
We create a test set for New York City and Paris, comprised of 1292 pairs of text prompts and GPS tags in total. We attach two files {\tt nyc-eval.json} and {\tt paris-eval.json} to show the lists we use for each city. 

\subsection{Average images}\label{supp:avg}
Within a selected area, we have a set of sampled locations with GPS coordinates $\{\left(x_{0},y_{0}\right), \left(x_{1},y_{1}\right), ..., \left(x_{M-1},y_{M-1}\right)\}$, then we could get their corresponding GPS embeddings $\{\mathbf{g}_{0},...,\mathbf{g}_{M-1}\}$. For the concept like text prompt ``{\tt building}'', we obtain fixed text embedding $\mathbf{p}$ for $\{\mathbf{g}_{0},...,\mathbf{g}_{M-1}\}$. The noise estimate is as follows:
{\small \begin{align}\label{eq:avg}
\bar{\boldsymbol{\epsilon}}_{\phi}\left(\mathbf{z}_{t}; \mathbf{p}, \mathbf{g}, t\right) &= \boldsymbol{\epsilon}_{\phi}\left(\mathbf{z}_{t};\varnothing, \varnothing, t\right) \notag \\
    &+ \omega_{\mathbf{p}}\left(\boldsymbol{\epsilon}_{\phi}\left(\mathbf{z}_{t};\mathbf{p}, \varnothing, t\right) - \boldsymbol{\epsilon}_{\phi}\left(\mathbf{z}_{t};\varnothing, \varnothing, t\right)\right) \notag \\
    &+ \omega_{\mathbf{g}}\left(\frac{\sum_{i=0}^{M-1}\boldsymbol{\epsilon}_{\phi}\left(\mathbf{z}_{t};\mathbf{p},\mathbf{g}_{i}, t\right)}{M} - \boldsymbol{\epsilon}_{\phi}\left(\mathbf{z}_{t};\mathbf{p}, \varnothing, t\right)\right)\text{,} 
\end{align}}where $\omega_{\mathbf{p}}$ and $\omega_{\mathbf{g}}$ are guidance weights also used in~\cref{dual_inference}, and $\boldsymbol{\epsilon}_{\phi}$ is the denoiser of our trained GPS-to-image diffusion model.
It is worth noting that all average images shown in~\cref{fig:average} share the same initial random noise. 

\subsection{GPS-CLIP}\label{supp:gps-clip}
As mentioned in~\cref{sec:experiments}, we use GPS score which measures cosine similarity between image and GPS embeddings as one of evaluation metrics. Specifically, we use pretrained frozen DINOv2 (ViT-B/14)~\cite{oquab2023dinov2} as image encoder. We add a single projection layer to the image encoder. For the GPS encoder, we use a shared-weight 6-layer MLP for latitude and longitude. The resulting embeddings are concatenated and passed through a single layer to produce the final GPS embedding, which has the same dimensionality as DINOv2. We use GELU~\cite{hendrycks2016gaussian} as activation function for GPS encoder. The batch size is 512, and temperature is 0.07, learning rate is $1\times10^{-4}$ with warmup and cosine learning rate decay~\cite{loshchilov2016sgdr}. We train GPS-CLIP on a single  NVIDIA L40S. The pseudocode for training process is presented in~\cref{alg:gps-clip}. 
\begin{algorithm}[h]
\caption{Pseudocode of training GPS-CLIP.}
\label{alg:gps-clip}
\algcomment{\fontsize{7.2pt}{0em}\selectfont 
\texttt{mm}: matrix multiplication.
}
\definecolor{codeblue}{rgb}{0.25,0.5,0.5}
\lstset{
  backgroundcolor=\color{white},
  basicstyle=\fontsize{7.2pt}{7.2pt}\ttfamily\selectfont,
  columns=fullflexible,
  breaklines=true,
  captionpos=b,
  commentstyle=\fontsize{7.2pt}{7.2pt}\color{codeblue},
  keywordstyle=\fontsize{7.2pt}{7.2pt},
}
\begin{lstlisting}[language=python]
# x, y: batch of longitudes and latitudes 
# imgs: batch of images
# f_gps: shared-weight encoder for longitude and latitude
# f_v: vision encoder of DINOv2
# p: projection layer for f_gps
# q: projection layer for f_v
#  t: temperature
for imgs, x, y in loader: # load a minibatch 
    x_f = f_gps.forward(x) 
    y_f = f_gps.forward(y)
    gps_e = p.forward(cat([x_f, y_f], dim=1)) # GPS embedding
    img_f = f_v.forward(imgs) 
    img_e = q.forward(img_f) # image embedding
    gps_e = gps_e / norm(gps_e) # embedding normalization
    img_e = img_e / norm(img_e) # embedding normalization
    logits = mm(img_e.view(1, C), gps_e.view(1, C).T)/t  
    labels = torch.arange(n)
    loss_i = cross_entropy_loss(logits, labels, axis=0)
    loss_g = cross_entropy_loss(logits, labels, axis=1)
    loss = (loss_i + loss_g)/2
    loss.backward()
\end{lstlisting}
\end{algorithm}

\subsection{Implementation details}\label{supp:gps-to-image-details}
We use \href{https://huggingface.co/Salesforce/xgen-mm-phi3-mini-instruct-r-v1}{xgen-mm-phi3-mini-instruct-r-v1} of BLIP-3~\cite{xue2024xgen} as our captioning model for collected datasets. For classifier-free guidance (CFG), we set $\omega_{\mathbf{p}}$ to 3.5 and $\omega_{\mathbf{g}}$ to 7.5 in~\cref{dual_inference} for GPS-to-image diffusion.

\begin{figure*}[!t]
    \centering
    \vspace{-2mm}
     \includegraphics[width=\linewidth]{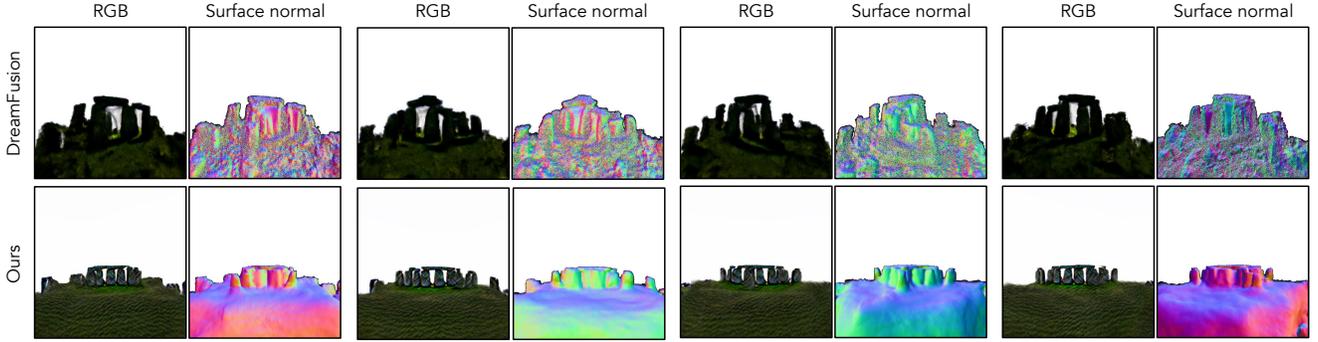}
    \caption{{\bf More qualitative comparison for 3D monument reconstruction.} We show qualitative results of DreamFusion~\cite{poole2023dreamfusion} and our method on Stonehenge. Our reconstructed 3D monuments have better visual quality and more accurate 3D structure. We use rendered depth to make the background of RGB rendering white. {\bf Please see \href{https://cfeng16.github.io/gps-gen/}{project webpage} for more examples.}} 
    \label{fig:3d_recon_supp}
\end{figure*}

\section{GPS-guided 3D reconstruction}
\begin{itemize}
    \item More 3D qualitative comparisons between our method and DreamFusion~\cite{poole2023dreamfusion} are presented in~\cref{fig:3d_recon_supp} and \href{https://cfeng16.github.io/gps-gen/}{project webpage}. Please refer to~\cref{3d_supp} for more details.
    \item Qualitative results regarding SfM~\cite{schonberger2016structure}, Nerfacto~\cite{tancik2023nerfstudio}, and NeRF-W~\cite{martinbrualla2020nerfw} are shown in~\cref{fig:sfm-nerf}. Please refer to~\cref{subsec:nerf-sfm} for more details.
    \item Some qualitative results of angle-to-image diffusion are presented in~\cref{fig:diffusion_qualitative}, please refer to~\cref{subsec:angle_supp} for more details.
\end{itemize}
\subsection{Angle-to-image diffusion}\label{subsec:angle_supp}
\begin{figure*}[ht]
    \centering
     \includegraphics[width=\linewidth]{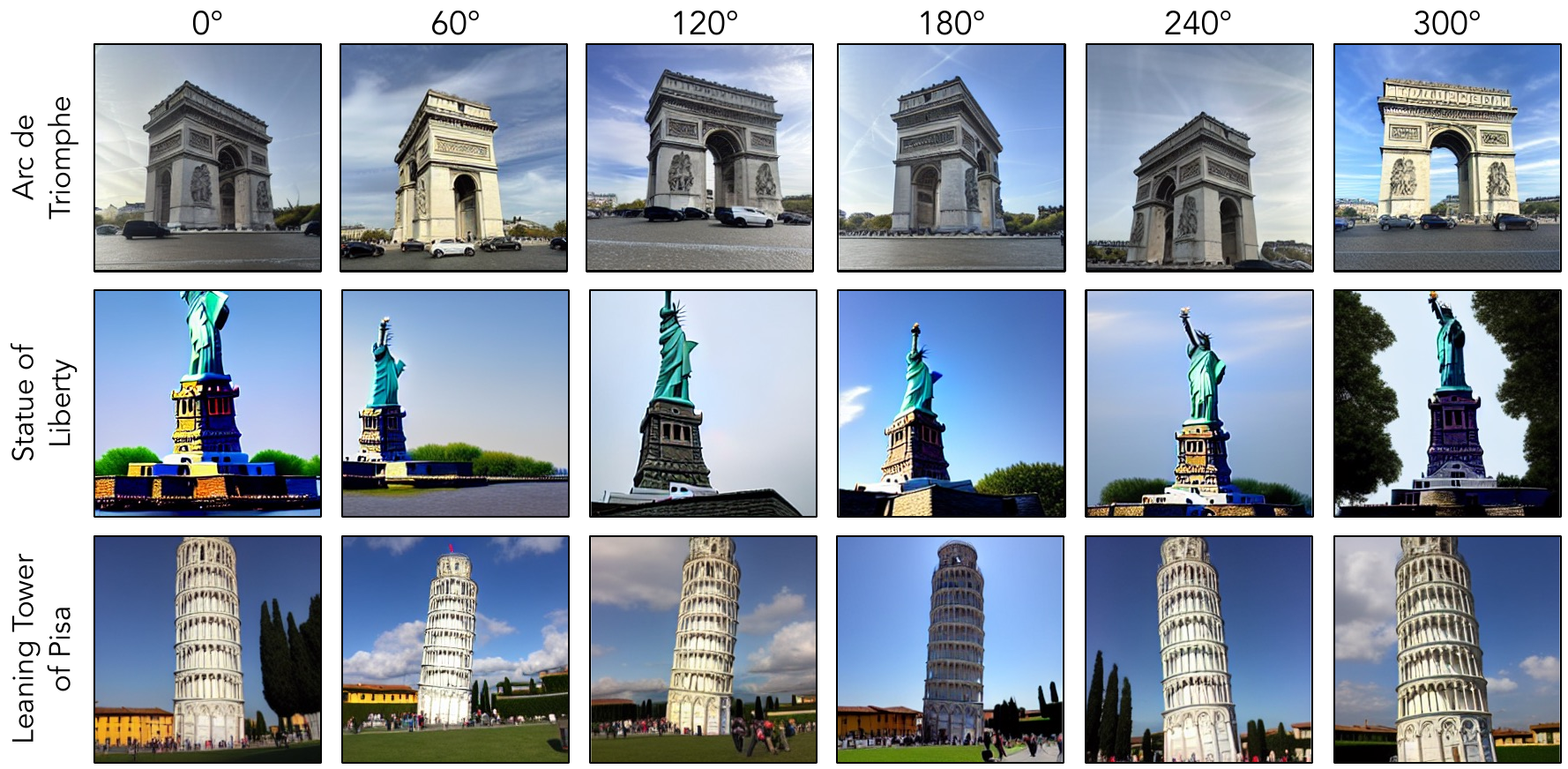}
    \caption{{\bf Qualitative results for angle-to-image generation.} We show generated images of our angle-to-image diffusion model for the Arc de Triomphe, Statue of Liberty, and Leaning Tower of Pisa. Images are sampled conditioned on different angles estimated by GPS tags.} 
    \label{fig:diffusion_qualitative}
\end{figure*}

\paragraph{Prior preservation loss}
As mentioned in~\cref{subsec:3d}, for angle-to-image model training, we utilize prior preservation loss. To be specific, with synthesized images $\mathcal{X^*}$ from original stable diffusion model~\cite{rombach2022high} and text condition $\mathbf{p}$, we optimize the preservation loss: 
\begin{equation}\label{prese_loss}
    \mathcal{L}_{\mathrm{preservation}} = \mathbb{E}_{\mathbf{x^*}, \boldsymbol{\epsilon}, t} \left[ \parallel \boldsymbol{\epsilon}_{t} - \boldsymbol{\epsilon}_{\phi}(\mathbf{z}_{t}^*; \mathbf{p}, \varnothing, t) \parallel_{2}^{2}\right] \text{,}
\end{equation} where $\varnothing$ represents that we zero out the angle condition for these training examples.
\paragraph{Implementation details}\label{supp:angle-to-image details} For each landmark, we finetune Stable Diffusion-v1.4~\cite{rombach2022high} on collected Flickr images, at a resolution of $256\times256$ for 800 steps. After angle discretization, we normalize the angle to the range of $[-1, 1]$. We use a positional encoding and a two-layer MLP to encode the angle condition. For the positional encoding, we use 10 frequencies. We use the AdamW~\cite{loshchilov2017decoupled} optimizer with a constant learning rate of $5\times10^{-6}$ and gradient accumulation without warm-up. We use a global batch size of 256 on 4 NVIDIA A40 GPUs.
\paragraph{Qualitative results}\label{supp:angle-to-image results}Some generated images from our angle-to-image diffusion model are presented in \cref{fig:diffusion_qualitative}.
\subsection{GPS-guided score distillation sampling}\label{3d_supp}
\paragraph{Gradient} The gradient in~\cref{subsec:3d} we use to supervise NeRF is as follows:
\begin{align}\label{SDS}
    \nabla_{\mathbf{\theta}}\mathcal{L}_{SDS}\left(\phi, \mathbf{x}=h_\theta\left(\mathbf{q}\right)\right) \approx  \notag \\
    \mathbb{E}_{\mathbf{g}\prime, \boldsymbol{\epsilon}_{t}, t} \left[\omega\left(t\right)\left(\hat{\boldsymbol{\epsilon}}_{\phi}\left(\mathbf{z}_{t}; \mathbf{p}, \mathbf{g}\prime, t\right) - \boldsymbol{\epsilon}_{t}\right)\frac{\partial \mathbf{x}}{\partial \mathbf{\theta}}\right] \text{,}
\end{align} 
where $\omega\left(t\right)$ is a weighting function, which we set to $\omega\left(t\right) = \sigma^{2}_{t}$ following~\cite{poole2023dreamfusion}.
\paragraph{Qualitative results}\label{supp:gps-to-3d results}
We show more qualitative comparison between our method and DreamFusion~\cite{poole2023dreamfusion} in~\cref{fig:3d_recon_supp} and \href{https://cfeng16.github.io/gps-gen/}{project webpage}. It should be noted that for all videos, we directly use {\bf raw} renderings and do {\bf not} use rendered depth to make the background of RGB rendering white.
\subsection{Baseline of COLMAP with NeRF}\label{subsec:nerf-sfm}
\begin{figure*}[!t]
    \centering
     \includegraphics[width=\linewidth]{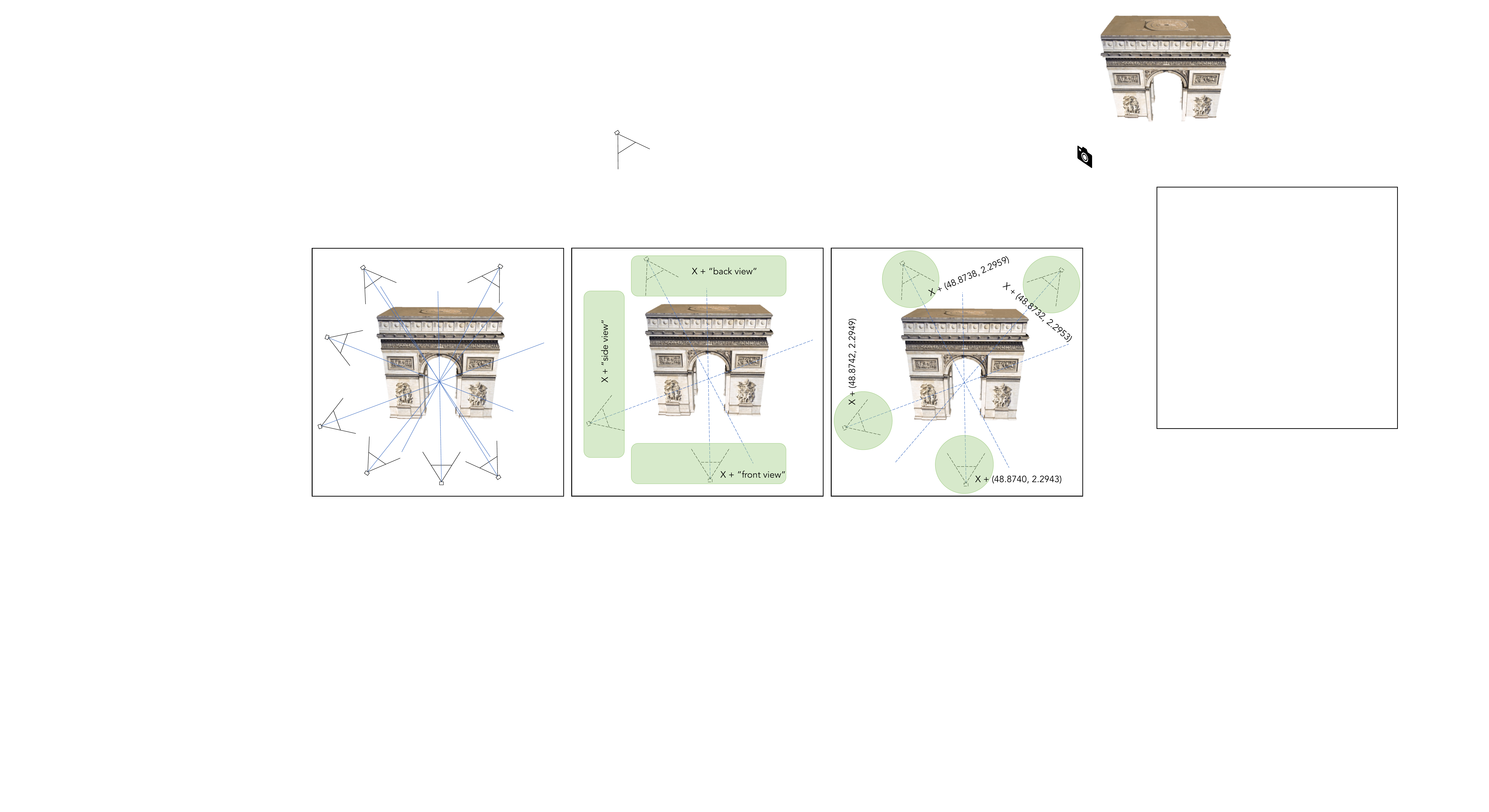}
   \caption{{\bf SfM/NeRF baselines.} We present SfM reconstructions from COLMAP~\cite{schoenberger2016sfm}, Nerfacto~\cite{tancik2023nerfstudio} rendering results, and NeRF-W~\cite{martinbrualla2020nerfw} rendering results for 6 evaluated landmarks. SfM reconstruction fails on (a), (b), and (c). Nerfacto~\cite{tancik2023nerfstudio} only succeeds on (f). NeRF-W~\cite{martinbrualla2020nerfw} completely fails on 6 scenes.}
    \label{fig:sfm-nerf}
\end{figure*}

We present qualitative results of COLMAP~\cite{schonberger2016structure}, NeRF-W~\cite{martinbrualla2020nerfw}, and Nerfacto~\cite{tancik2023nerfstudio} in~\cref{fig:sfm-nerf}. Since NeRF-W's~\cite{martinbrualla2020nerfw} official code is not available, we evaluate the popular \href{https://github.com/kwea123/nerf_pl/tree/nerfw}{reimplementation}. As shown in~\cref{fig:sfm-nerf}, COLMAP~\cite{schoenberger2016sfm} successfully reconstructs camera poses and sparse point clouds for 3 of the 6 scenes, and fails on 3. NeRF-W~\cite{martin2021nerf} estimation completely fails on 6 landmarks and Nerfacto~\cite{tancik2023nerfstudio} fails on 5.

\section{Datasets}
\begin{figure*}[ht]
    \centering
     \includegraphics[width=\linewidth]{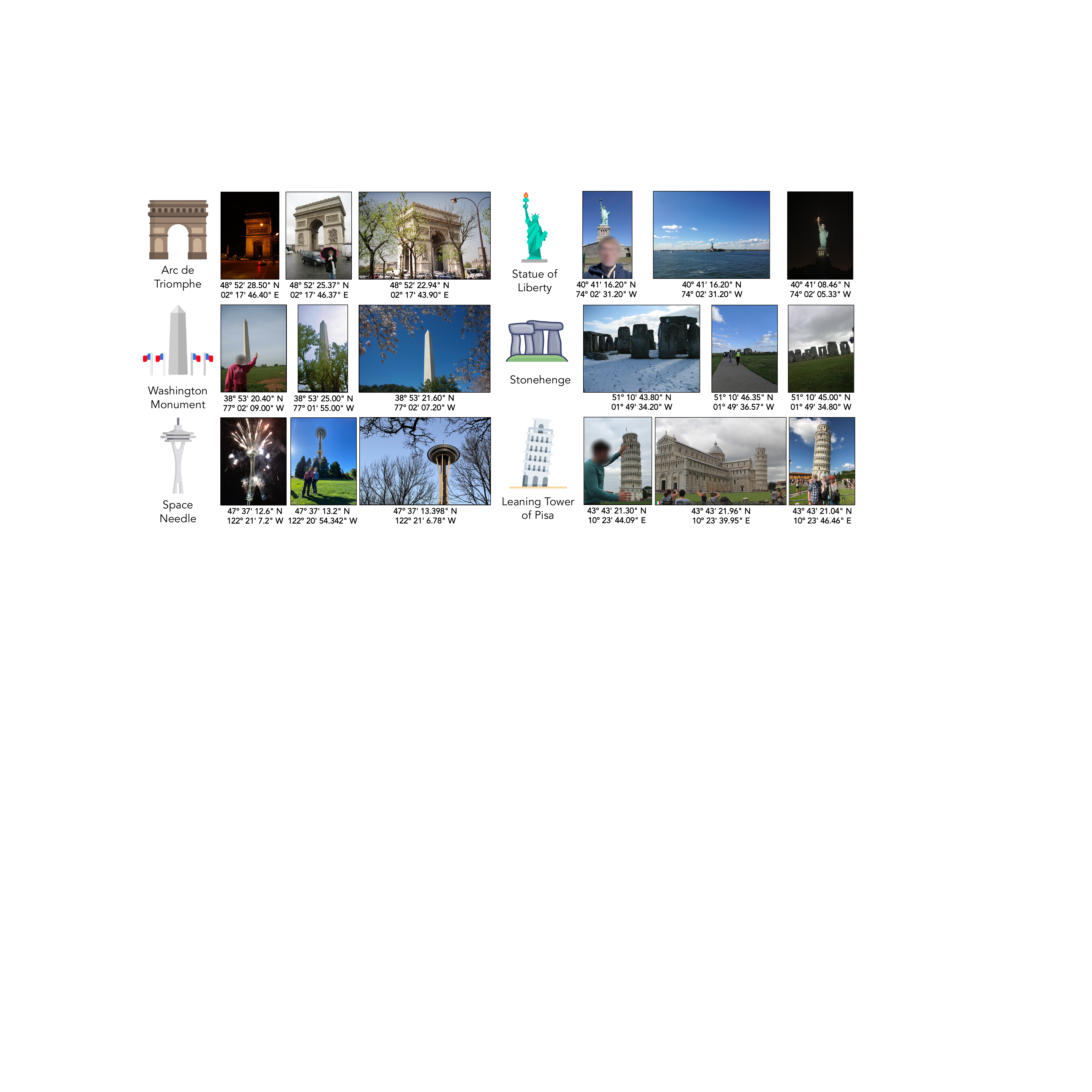}
   \caption{{\bf Data samples.} We show some random photos with their GPS tags from our collected datasets.}
    \label{fig:data_sample}
\end{figure*}

\paragraph{Tourist photo collection}\label{supp:dataset}By querying Flickr, we obtain photo collections for 2 cities:  \textbf{1)} New York City (Manhattan, 501,592 photos); \textbf{2)} Paris (315,306 photos) and 6 landmarks: \textbf{1)} Leaning Tower of Pisa (2,967 photos); \textbf{2)} Arc de Triomphe (2,377 photos); \textbf{3)} Washington Monument (2,563 photos); \textbf{4)} Statue of Liberty (1,174 photos); \textbf{5)} Stonehenge (2,486 photos); \textbf{6)} Space Needle (1,800 photos). The number of evaluated landmarks is {\em in-line with prior work}~\cite{martinbrualla2020nerfw} in the field. It is worth noting that we focus primarily on {\bf Manhattan} for New York City due to resource constraints. Some examples sampled from datasets are shown in~\cref{fig:data_sample}. It should be noted that 2 cities are collected for GPS-to-image generation and 6 landmarks are for angle-to-image generation and 3D landmark reconstruction. As mentioned in~\cref{subsec:3d}, the angle of capture is necessary so we use a bespoke dataset for each landmark.

\end{document}